\pdfoutput=1

\documentclass[11pt]{article}

\usepackage{acl}

\usepackage{times}
\usepackage{bm}
\usepackage{latexsym}

\usepackage[T1]{fontenc}

\usepackage[utf8]{inputenc}
\usepackage{amsmath}
\usepackage{microtype}

\usepackage{inconsolata}

\usepackage{graphicx}
\usepackage{makecell}

\usepackage{amsfonts}
\usepackage{booktabs}
\usepackage{multirow}
\usepackage{booktabs, multirow, array}

\title{RAG-Zeval: Towards Robust and Interpretable Evaluation on RAG Responses through End-to-End Rule-Guided Reasoning}

\author{Kun Li$^{\heartsuit}$\thanks{$\;\;$Equal contribution.}$\,$, Yunxiang Li$^{\heartsuit*}$, Tianhua Zhang$^{\heartsuit*}$, Hongyin Luo$^{\diamondsuit}$, \\ \bf
Xixin Wu$^{\heartsuit}$, James Glass$^{\diamondsuit}$, Helen Meng$^{\heartsuit}$ \\
$^\heartsuit$The Chinese University of Hong Kong, Hong Kong SAR, China \\
$^\diamondsuit$Massachusetts Institute of Technology, Cambridge MA, USA \\
\texttt{\{li.kun, yli, thzhang\}@link.cuhk.edu.hk}
}

\begin{document}
\maketitle
\begin{abstract}

Robust evaluation is critical for deploying trustworthy retrieval-augmented generation (RAG) systems. However, current LLM-based evaluation frameworks predominantly rely on directly prompting resource-intensive models with complex multi-stage prompts, underutilizing models' reasoning capabilities and introducing significant computational cost. In this paper, we present \texttt{RAG-Zeval} (\textbf{RAG}-\textbf{Z}ero \textbf{Eval}uator), a novel end-to-end framework that formulates faithfulness and correctness evaluation as a rule-guided reasoning task. Our approach trains evaluators with reinforcement learning,
facilitating compact models to generate comprehensive and sound assessments with detailed explanation in one-pass. We introduce a ranking-based outcome reward mechanism, using preference judgments rather than absolute scores, to address the challenge of obtaining precise pointwise reward signals. To this end, we synthesize the ranking references by generating quality-controlled responses with \textit{zero} human annotation.
Experiments demonstrate \texttt{RAG-Zeval}'s superior performance, achieving the strongest correlation with human judgments and outperforming baselines that rely on LLMs with $10-100\times$ more parameters. Our approach also exhibits superior interpretability in response evaluation.

\end{abstract}
\section{Introduction}

Retrieval-Augmented Generation (RAG) systems \citep{lewis2021retrievalaugmentedgenerationknowledgeintensivenlp, gao2024retrievalaugmentedgenerationlargelanguage, li2024decodinggraphsfaithfulsound} have become a cornerstone for building knowledge-intensive NLP applications, such as question answering, fact-checking in various domains. \cite{zhao2025medrag, Pipitone2024LegalBenchRAGAB} By integrating external knowledge retrieval with large language models (LLMs), RAG enables more accurate and contextually relevant responses \citep{li2025generatediscriminateevolveenhancing, asai2024selfrag}, especially for queries that go beyond the static knowledge encoded in model parameters. As RAG systems are increasingly deployed in real-world scenarios, robust and comprehensive evaluation is essential to assess their performance and guide further development \citep{Yu_2025}.

However, evaluating RAG systems remains challenging due to their modular structure and the complex interplay between retrieval and generation. Traditional metrics such as recall@k and MRR for retrieval, and BLEU \cite{papineni-etal-2002-bleu}, and ROUGE \cite{lin-2004-rouge} for generation are often coarse-grained and fail to capture semantic fidelity or factual consistency in open-ended tasks. To overcome these limitations, recent work has explored model-based evaluation strategies, particularly leveraging LLMs as automatic judges \cite{Gu2024ASO}. Frameworks such as RAGAS \cite{Shahul2023RAGAsAE} and RAG-Checker \cite{Ru2024} have demonstrated that LLMs can provide scalable and automated assessments of metrics like context relevance and faithfulness, reducing the reliance on costly human annotation and enabling efficient large-scale evaluation.

Although showing the potential as being automated RAG evaluators, these LLM-based approaches \citep{Shahul2023RAGAsAE,Ru2024} predominantly rely on prompting large-scale LLMs with advanced capabilities (e.g., GPT-4 \citep{openai2024gpt4technicalreport}, Llama3-70B \citep{grattafiori2024llama3herdmodels}) to achieve strong assessment results, introducing significant computational costs. Motivated by these limitations, we study whether \textit{compact LLMs can be transformed into end-to-end and interpretable evaluators with incentivized reasoning abilities.}

In this work, we present \texttt{RAG-Zeval} (\textbf{RAG}-\textbf{Z}ero \textbf{Eval}uator), a novel framework that formulates faithfulness and correctness evaluation as a rule-guided reasoning task  with zero human annotation. Our approach enables the compact evaluators to generate comprehensive and sound assessments in one-pass under the instruction of predefined rules, systematically performing (1) claim decomposition, (2) evidence grounding, and (3) supportiveness judgment. This \textit{end-to-end} evaluation distinguish our method from previous multi-stage pipelines, which ensures assessment consistency through holistic reasoning and captures the interdependence between different steps. Drawing inspiration from recent advances in reinforcement learning (RL) \citep{deepseekai2025deepseekr1incentivizingreasoningcapability,yu2025dapoopensourcellmreinforcement,liu2025understandingr1zeroliketrainingcritical}, we train evaluators using rule-based RL \citep{deepseekai2025deepseekr1incentivizingreasoningcapability} to elicit interpretable reasoning capabilities from compact LLMs. This rule-based RL alignment eliminates the need for process rewards (e.g., evaluation trajectory annotations) \citep{lightman2024lets, zhong2025comprehensivesurveyrewardmodels}. To overcome the challenge of acquiring precise point-wise reward signals, we introduce a \textit{ranking-based outcome reward} mechanism, which operates on more easily obtainable preference judgments instead of absolute scores \citep{guan2025rstarmathsmallllmsmaster}. Recognizing that high-quality rewards in open-ended generation tasks typically require expensive human annotations \citep{liu2025inferencetimescalinggeneralistreward}, we further synthesize
the ranking reference using Context-Aware Decoding \cite{Shi2023TrustingYE} to generate quality-controlled response candidates. This enables fully automated training free of human labels.
Additionally, we employ curriculum learning, progressively increasing the number of candidate responses to be ranked as RL advances, which further improves the evaluators performance.

We assess \texttt{RAG-Zeval} on both faithfulness and correctness benchmarks to analyze its performance in deriving interpretable and reliable evaluations. Experimental results demonstrate that \texttt{RAG-Zeval} achieves strong alignment with human judgments, maintaining transparent and interpretable decision-making through its rule-guided reasoning process.


\section{Related Work}
With the rapid advancement of Retrieval-Augmented Generation (RAG) systems \citep{fan2024surveyragmeetingllms, li2025generatediscriminateevolveenhancing}, effective and robust evaluation methods beyond traditional metrics have become increasingly important. 

A significant line of work evaluates the retrieval and generation components separately. For retrieval, traditional information retrieval metrics such as precision, recall, MRR, and MAP are widely used. \cite{Yu2024EvaluationOR,tang2024multihoprag}  For the generation component, metrics like BLEU \cite{papineni-etal-2002-bleu}, ROUGE \cite{lin-2004-rouge}, and BERTScore \cite{bert-score} are commonly used, alongside human evaluation. Such component-wise evaluation often fails to capture the complex interactions between retrieval and generation in real-world RAG systems, and human annotation is costly, time-consuming, and subject to annotator bias and inconsistency.

Recent research on the evaluation of RAG systems has moved beyond traditional component-wise metrics, proposing a variety of end-to-end frameworks that leverage large language models (LLMs) as evaluators. LLM-based evaluation frameworks such as TruLens \cite{ferrara2024ragtriad} and ARES \cite{saad-falcon-etal-2024-ares} adopt direct prompting to score responses without decomposing them into individual claims. Other approaches, including RAGAS \cite{Shahul2023RAGAsAE}, RAG-Checker \cite{Ru2024}, and OpenEval \cite{Ispas2025TowardsLA}, introduce claim-level decomposition, enabling LLMs to assess the faithfulness and correctness of each factual statement for finer-grained and more interpretable evaluation.


Despite these advances, most current LLM-based evaluation frameworks rely on direct prompting of large, resource-intensive models, often involve complex multi-stage prompting, and treat LLMs as black-box scorers without fully leveraging their reasoning abilities. Recent progress in reinforcement learning and reward modeling, such as Deepseek-R1 \cite{deepseekai2025deepseekr1incentivizingreasoningcapability} and generative reward modeling (GRM) \citep{liu2025inferencetimescalinggeneralistreward}, demonstrates that rule-driven, interpretable evaluators trained via rule-based RL can provide more transparent and scalable assessments with stronger reasoning ability. These developments motivate our approach to construct RAG evaluators using similar RL-based, rule-guided techniques.
\section{Methodology}
\begin{figure*}[ht]
\centering
\includegraphics[width=0.9\textwidth]{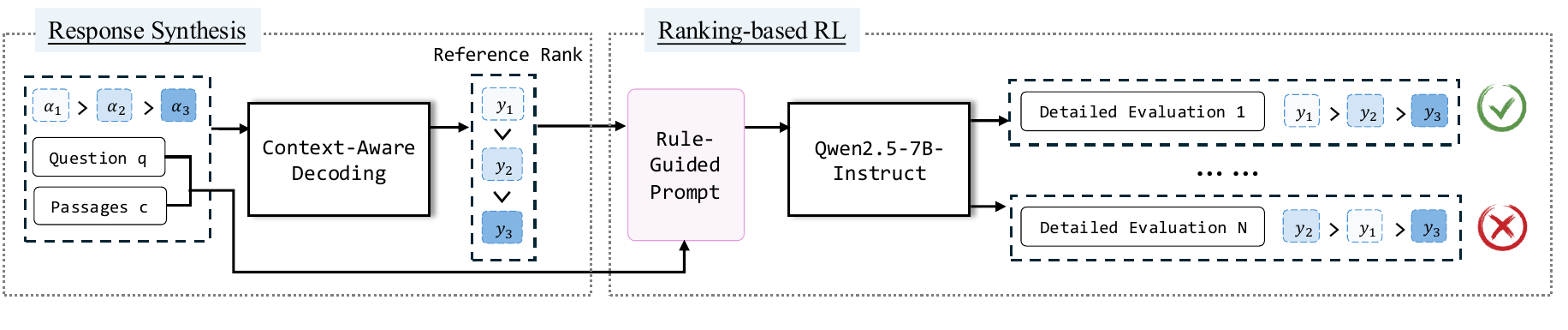}
\caption{An overview of \texttt{RAG-ZEval}. We synthesize training data using Context-Aware Decoding. The complete prompt is presented in Fig.\ref{fig:prompt}. The ground-truth ranking of $\bm{y}$'s depends on the value of $\alpha$.}
\label{fig:diagram}
\vspace{-0.5cm}
\end{figure*}

\subsection{Problem Formulation}
A majority of evaluation tools for assessing the response quality of RAG systems adopt a claim-based paradigm\cite{Ru2024, Shahul2023RAGAsAE, Ispas2025TowardsLA, manakul-etal-2023-selfcheckgpt}. In this paradigm, system responses are decomposed into individual claims, a declarative sentence that conveys an atomic piece of information, and each claim is then evaluated for supportiveness—the degree to which it is grounded in the provided reference context (e.g., ground-truth answer for correctness or retrieved passages for faithfulness). 

In a RAG setting, a response is considered 1) correct if the ground-truth answer supports the claims of the response, or 2) faithful if the retrieved passage supports the claims of the response. Based on this, following prior work \cite{Ru2024, Ispas2025TowardsLA, Shahul2023RAGAsAE}, given a response, we define its \textbf{\textit{correctness}} as
\begin{equation}
\resizebox{0.6\linewidth}{!}{%
    $\frac{\text{\# Claims supported by ground-truth answer}}{\text{\# Claims}},$}
\label{eq:correctness}
\end{equation}
and \textbf{\textit{faithfulness}} as
\begin{equation}
\resizebox{0.64\linewidth}{!}{%
    $\frac{\text{\# Claims supported by retrieved passage}}{\text{\# Claims}},$}
\label{eq:faithfulness}
\end{equation}
indicating the precision rate of claims in the response that are supported by the ground-truth answer and the retrieved passage respectively. \textit{These formulations consolidate the evaluation paradigms for both correctness and faithfulness, allowing for the development of a unified evaluator that assesses the two quality dimensions}.
We can conduct evaluation in both correctness and faithfulness with different reference (ground-truth answer for correctness and retrieved passages for faithfulness evaluation).

\subsection{Prompting for Rule-Guided Reasoning}
\label{sec:prompt}
 
Different from previous claimed-based work which runs in a multi-stage pipeline, we develop a novel approach for end-to-end claim-based evaluation, through \textit{generation} of complete evaluation trajectories under the rules of guide. 

To this end, we adopt the prompt demonstrated in Fig.\ref{fig:prompt}, which elaborates the rules and format the generation should conform to. In detail, given a question $\bm{q}$ and the reference $\bm{c}$, and the set of responses to evaluate $\{\bm{y}\}$, LLM should give a sound evaluation process---decomposing a response into claims, and then determining those claims' supportiveness as well as finding the grounding evidence in the reference. In addition, the generation is required to represent the evaluation process in a JSON format. After parsing the generated JSON-formatted string into a Python list object using \texttt{json.loads()}, we can easily extract the intermediate results (e.g., \# claims (un)supported by reference) by querying the resulting list object. 

Casting the evaluation process into generation of evaluation trajectory not only streamlines the pipeline, but also facilitates further finetuning of the model.

\subsection{Reinforcement Learning with Ranking Objective}

Finetuning models with valid evaluation trajectories as outlined in \S\ref{sec:prompt}, presents a non-trivial challenge due to the prohibitive cost of manual annotation---particularly for claim decomposition and supportiveness judgment. To address this, we use rule-based reinforcement learning \cite{deepseekai2025deepseekr1incentivizingreasoningcapability} to finetune our model, without the need for annotation of the whole trajectory. 

Nonetheless, constructing labeled data for reward calculation remains necessary. A naive way would be to annotate the score according to Eq.\ref{eq:correctness} or \ref{eq:faithfulness}. However, this way still relies on the claim decomposition and supportiveness judgment as the intermediate results. To circumvent this, our rule-based reinforcement learning method introduces a novel optimization paradigm that trains the model to perform relative ranking of candidate responses based on their degree of supportiveness w.r.t the reference, rather than predicting absolute scores. Specifically, first, given question $\bm{q}$ and reference $\bm{c}$, we synthesize a set of responses $\{\bm{y}\}$ with varying groundness degree w.r.t $\bm{c}$ (\S \ref{sec:pseudo_label_construction}). During this process, the ground-truth rank of $\{\bm{y}\}$ can be obtained naturally. Subsequently, based on $\bm{q}$ and $\bm{c}$, we adopt $\{\bm{y}\}$ as the candidates to rank and apply rule-based RL to reinforce the model's ranking ability by advancing the generated evaluation trajectories (\S \ref{sec:rule_based_rl}). Note that the RL objective is to rank the responses instead of predicting their exact scores. This can mitigate the adverse effects of bias introduced during data synthesis on the training.

\subsubsection{Responses Synthesis with Ranking Relation}
\label{sec:pseudo_label_construction}
With Context-Aware Decoding \cite{Li2022ContrastiveDO, Shi2023TrustingYE} , the $i$-th token of a response $y$ is sampled as
\begin{equation}
\begin{aligned}
    y_i \sim softmax[&(1+\alpha)P_{LLM} (*\mid \bm{q}, \bm{c}, \bm{y_{<i}}) \\
    &- \alpha P_{LLM} (*\mid \bm{q}, \bm{y_{<i}})].
\end{aligned}
\label{eq:cad}
\end{equation}
The weight $\alpha$ controls the extent in which the generation of $y_i$ is conditioned on \bm{$c$} (which is the passage in this case), and a larger one translates into \bm{$r$} that is more reference-conditioned. Note that $\alpha$ can be negative ($\alpha<0$), which leads to reference-resistant generation of \bm{$y$}.   

For each question, we synthesize a set of responses $\{\bm{y_{a_i}}\}$ with different degrees of groundness by varying $\alpha$. The ground-truth ranking of these responses can be obtained naturally as
\begin{equation}
\forall \alpha_i, \alpha_j \in \mathbb{R}, \quad \alpha_i > \alpha_j \implies \bm{y}_{\alpha_i} \succ \bm{y}_{\alpha_j}.
\end{equation}

For implementation, $P_{LLM} (*\mid \bm{q}, \bm{c}, \bm{r_{<i}})$ and $P_{LLM} (*\mid \bm{q}, \bm{r_{<i}})$ are modeled using in-context learning. We use a third-party LLM for sampling candidate responses prior to the RL stage.

\subsubsection{Rule-Based RL}
\label{sec:rule_based_rl}
We fine-tune the model with rule-based  RL. Particularly, we adopt Group Relative Policy Optimization (GRPO, \citeauthor{Shao2024DeepSeekMathPT}, \citeyear{Shao2024DeepSeekMathPT}) with rule-based outcome rewards. During rolling out, with the question $\bm{q}$, the reference $\bm{c}$, and the set of synthesized responses $\{\bm{y_{a_i}}\}$ as input, the model generates complete evaluation trajectories according to the rules specified in the prompt.

\vspace{10pt}
\noindent\textbf{Reward Design} We define three types of rewards, including format reward, evidence reward, and accuracy reward. The rewards for the rollout of evaluation trajectory are defined as follows. 
\begin{itemize}
\item \textbf{Format reward} assesses the completeness of the evaluation trajectory. $r_\text{f}$ is $\mathbf{0}$ if the string of evaluation trajectory satisfies all the following requirements: 1) it can be parsed into a Python \texttt{List} object using \texttt{json.loads()}; 2) the items in the list correspond exactly to the set of candidate responses; 3) each item within the \texttt{List} is a \texttt{Dict} object containing all required fields as specified in the prompt (the circled region in Fig.\ref{fig:prompt}); 4) each supported claim has at least one evidence. Otherwise a penalty of $\mathbf{-0.5}$ is applied.

\item \textbf{Evidence reward} measures how verbatim extracted evidence spans are cited from the reference. The reward for each span is defined as the length of its longest common substring with the reference text, normalized by the span's length\footnote{The length of a sequence is computed as its total token count}. An evidence span of length less than 10 receives reward $\mathbf{0}$. The evidence reward of an evaluation trajectory $r_\text{e}$ is the average over all evidence spans in the trajectory.

\item \textbf{Accuracy reward} evaluates whether the ranking based on evaluation scores inferred by the model is correct. The evaluation score is derived as $S(\bm{y})=\frac{\text{\# Claims of } \bm{y} \text{ supported by }\bm{c}}{\text{\# Claims of }\bm{y}}$.The accuracy reward $r_\text{a}$ is $\mathbf{1}$ if the ranking aligns with the ground-truth ranking, or $\mathbf{0}$ otherwise. Formally,
\end{itemize}
\begin{equation}
\resizebox{1.035\linewidth}{!}{%
    $r_\text{a} = \left\{
    \begin{array}{ll}
    1,  \,\text{if } S(\bm{y}_{\alpha_i}) > S(\bm{y}_{\alpha_j}), \forall&\!\!\!\bm{y}_{\alpha_i}, \bm{y}_{\alpha_j} \in \{\bm{y}\}\, \text{and}\, \\
    &\!\!\!\bm{y}_{\alpha_i} \succ \bm{y}_{\alpha_j}\\
    0,  \,\text{otherwise}
    \end{array}
    \right.$}
\label{eq:acc_reward}
\end{equation}

\noindent Note that the intermediate results required for reward calculation can be accessed by visiting the object parsed using \texttt{json.loads()}. For instance, for each candidate response, we apply  simple Python operations to enumerate all entries in its \texttt{atomic\_claims} list and verify their \texttt{is\_supported} values. The JSON-formatted output demonstrates superior precision in results extraction, compared to traditional regular expression-based approaches.

\noindent Taking together above three rewards, the combined reward $r$ for a rollout is
\begin{equation}
r=\left\{\begin{array}{ll}
1+0.5*r_\text{e},  &\text{if}  \, r_\text{f}=0 \, \text{and} \, r_\text{a}=1,\\
0,  &\text{if} \, r_\text{f}=0 \, \text{and} \, r_\text{a}=0,\\
-0.5, & \text{otherwise}.
\end{array}\right.
\label{eq:final_reward}
\end{equation}

\vspace{-0.3cm}

The reward function encourages the model to rank the candidate responses more accurately through optimizing the evaluation trajectories. 

\vspace{10pt}
\noindent\textbf{Curriculum Learning} Intuitively, it is more challenging to rank a larger set of candidate responses. In the spirit of curriculum learning \cite{10.1145/1553374.1553380,10.5555/3455716.3455897}, to facilitate smooth and incremental learning, we gradually escalate the complexity of the ranking task by increasing the number of candidate responses as the RL training process advances.

\section{Experiment Settings}

\subsection{Benchmarks}
\label{sec:benchmarks}
\noindent \textbf{Faithfulness} We assess the faithfulness judgment performance of different evaluation approaches on WikiEval dataset \citep{Shahul2023RAGAsAE}, which contains question-context-answer triples with human-annotated judgments. The questions are formulated from 50 Wikipedia pages, and for each question, ChatGPT generates two answers: one with Wikipedia context and one without. Two human annotators then judge which answer is more faithful to the source, reaching 95\% agreement.

\noindent \textbf{Correctness} To assess different correctness evaluation approaches, we use the Meta Evaluation Dataset constructed by \citet{Ru2024}. The dataset contains 280 instances from 10 domains. Each instance includes a question, the ground-truth answer, and a pair of responses generated by two RAG systems\footnote{\url{https://github.com/amazon-science/RAGChecker/blob/main/data/meta_evaluation/}}. 
Two human annotators assess the responses, assigning preference labels from five relative choices: significantly better, slightly better, tie, slightly worse and significantly worse. We adopt human-annotated correctness preferences as the references to benchmark evaluation methods.

\subsection{Metrics}
\noindent \textbf{Faithfulness} For each WikiEval instance, the evaluators are required to identify the more faithful answer between two candidates. Evaluator performance is then measured as the percentage of cases where the evaluators' preference aligns with the human annotators' judgment \citep{Shahul2023RAGAsAE,Ispas2025TowardsLA}. We follow \citet{Shahul2023RAGAsAE} to handle possible \textit{ties} with three scenarios (see App. \ref{sec:metric-appendix} for more details): 
\begin{itemize}
\vspace{-0.1cm}
\setlength{\itemsep}{0pt}
    \item \textbf{Best-Case}: Measures the frequency of evaluators assigning greater or equal faithfulness scores to good answers over poor ones. 
    \item \textbf{Worst-Case}: Computes the frequency of strictly greater faithfulness scores assigned to good answers.
    \item \textbf{Middle-Case}: Adopts ternary scoring with a partial point of $0.5$ for ties.
\end{itemize}

\noindent \textbf{Correctness} Following \citet{Ru2024}, we covert the human-annotated correctness preference labels (five relative choices) into a numerical score difference for each response pair, i.e., $h_i=H(r_i^2)-H(r_i^1)\in\{-2,-1,0,1,2\}$. A normalized score difference is computed as $e_i=f(E(r_i^2)-E(r_i^1))$ for each evaluation approach, where $E(\cdot)$ is the correctness score measured by the evaluator and $f(\cdot)$ is a linear normalization function. To assess the performance of different evaluation methods, we compute three correlation coefficients between human judgments $h_i$ and system scores $e_i$: Pearson's $r$, Spearman's $\rho$, and Kendall's $\tau$.

\subsection{Implementation Details}
\textbf{Responses Synthesis} We use Natural Question dataset \cite{Kwiatkowski2019NaturalQA} to synthesize the responses, where each question is accompanied by a grounding passage. 5,500 instances are selected for response synthesis. For each $\alpha \in \{0, -0.5, -1, -1.4\}$, we synthesize a response according to Eq.\ref{eq:cad}, using \texttt{Qwen2.5-7B-Instruct} \cite{qwen2.5}. See App.~\ref{sec:synthesis-appendix} for more details. 

\noindent\textbf{Training} We fine-tune our model  based on \texttt{Qwen2.5-7B-Instruct}. For RL training, the sample number is 8 and temperature is 1 during rollout. The KL coefficient in the learning objective is 0.015. We train the model for a total of 2 epochs for training. To achieve curriculum learning for RL, we use 3 candidate responses for ranking in the first epoch and increase this to 4 in the second epoch. More details can be found in App.~\ref{sec:ours-implementation-appendix}.

\noindent\textbf{Inference} For test instances on both datasets described in \S\ref{sec:benchmarks}, similar to rolling out at training stage, the evaluator model takes as input the question, the reference text (ground-truth answer for correctness and retrieved passage for faithfulness), and two candidate responses; During the generation of evaluation trajectory, we use nucleus sampling \cite{Holtzman2019TheCC} with $p=0.9$ and temperature $=0.1$. For those generated sequence that fails to parse, we utilize regular expressions to extract the required results. The correctness/faithfulness score for a response $\bm{y}$ is computed as $S(\bm{y})$ (Eq.\ref{eq:acc_reward}, see Fig.\ref{fig:case_study_rl} for an example).

\begin{table}[t]
\scalebox{0.75}{
\begin{tabular}{llccc}
\toprule 
\textit{Method} & \textit{Model (-Instruct)} & \textit{Best} & \textit{Middle} & \textit{Worst} \\ \midrule \midrule
BLEU                                                                         & --        & 0.860 & 0.860  & 0.860 \\
RougL                                                                        & --        & 0.900 & 0.900  & 0.900 \\
BERTScore                                                                    & --        & 0.900 & 0.900  & 0.900 \\ \midrule
\multirow{3}{*}{ARES}                                                        & llama-70b & 1.000 & 0.920  & 0.840 \\
                                                                             & qwen-72b  & 1.000 & 0.928  & 0.856 \\
                                                                             & gpt-4o    & 1.000 & 0.956  & 0.912 \\ \midrule
\multirow{3}{*}{TruLens}                                                     & llama-70b & 1.000 & 0.860  & 0.720 \\
                                                                             & qwen-72b  & 0.984 & 0.830  & 0.676 \\
                                                                             & gpt-4o    & 1.000 & 0.940  & 0.900 \\ \midrule
\multirow{3}{*}{RAGAS}                                                       & llama-70b & 0.960 & 0.910  & 0.860 \\
                                                                             & qwen-72b  & 0.960 & 0.922  & 0.884 \\
                                                                             & gpt-4o    & 0.980 & 0.940  & 0.900 \\ \midrule
\multirow{3}{*}{RAG-Checker}                                                 & llama-70b & 1.000 & 0.962  & 0.924 \\
                                                                             & qwen-72b  & 0.976 & 0.936  & 0.896 \\
                                                                             & gpt-4o    & 0.973 & 0.933  & 0.893 \\ \midrule
OpenEval*                                                                         & llama-70b & 0.960 & 0.950  & 0.940 \\ \midrule
SFT                                                                          & qwen-72b  & 0.828 & 0.828  & 0.828 \\ \midrule
\multirow{3}{*}{\begin{tabular}[c]{@{}l@{}}RAG-Zeval \\ w/o RL\end{tabular}} & llama-70b & 0.980 & 0.960  & 0.927 \\
                                                                             & qwen-72b  & 0.993 & 0.957  & 0.883 \\
                                                                             & qwen-7b   & 0.932 & 0.930  & 0.858 \\ \midrule
RAG-Zeval                                                                    & qwen-7b   & $\textbf{1.000}^{\dagger}$ & $\textbf{0.992}^{\dagger}$  & $\textbf{0.984}^{\dagger}$ \\ \bottomrule \bottomrule
\end{tabular}}
\caption{Performance on \textbf{faithfulness} evaluation. We assess different methods using \texttt{Llama3.1-70B-Instruct}, \texttt{Qwen2.5-70B-Instruct}, \texttt{GPT-4o} and/or \texttt{Qwen2.5-7B-Instruct}. Non-\texttt{GPT} results are averaged over five trials to mitigate randomness. Due to API cost, we ran \texttt{GPT-4o} three times for each method. We cite results of OpenEval from the original paper \citep{Ispas2025TowardsLA}. $\dagger$ indicates the result is statistically significant at the level of 0.01.}
\vspace{-0.5cm}
\label{tab: faithfulness}
\end{table}

\subsection{Baselines}
We compare our approach with a comprehensive set of baseline evaluation methods, including non-LLM based and LLM-based paradigms. For non-LLM based methods, we report BLEU \cite{papineni-etal-2002-bleu} and ROUGE-L \cite{lin-2004-rouge} as representative n-gram based metrics, as well as BERTScore\cite{bert-score} for embedding-based metric. For LLM-based evaluation, we include recent frameworks that all use iterative prompting with large language models as evaluators. ARES \cite{saad-falcon-etal-2024-ares} and TruLens \cite{ferrara2024ragtriad} are non-claim-based, directly prompting the LLM for overall or aspect-based scores. RAGAS \cite{Shahul2023RAGAsAE}, RAG-Checker \cite{Ru2024}, and OpenEval \cite{Ispas2025TowardsLA} are claim-based, decomposing responses into factual claims for individual assessment. All LLM-based baselines use \texttt{Llama3.1-70B-Instruct,
Qwen2.5-70B-Instruct, GPT-4o} as the evaluator backbone.
In addition, we consider standard SFT, which directly fine-tunes \texttt{Qwen2.5-7B-Instruct} to replicate the relative ranking of responses, using the same synthetic data described in Section~\ref{sec:rule_based_rl} (see App.~\ref{sec:baseline-appendix} for more details). 

\section{Main Experiments}
\label{sec:main_experiment}
\begin{figure*}[ht]
\centering
\includegraphics[width=1\textwidth]{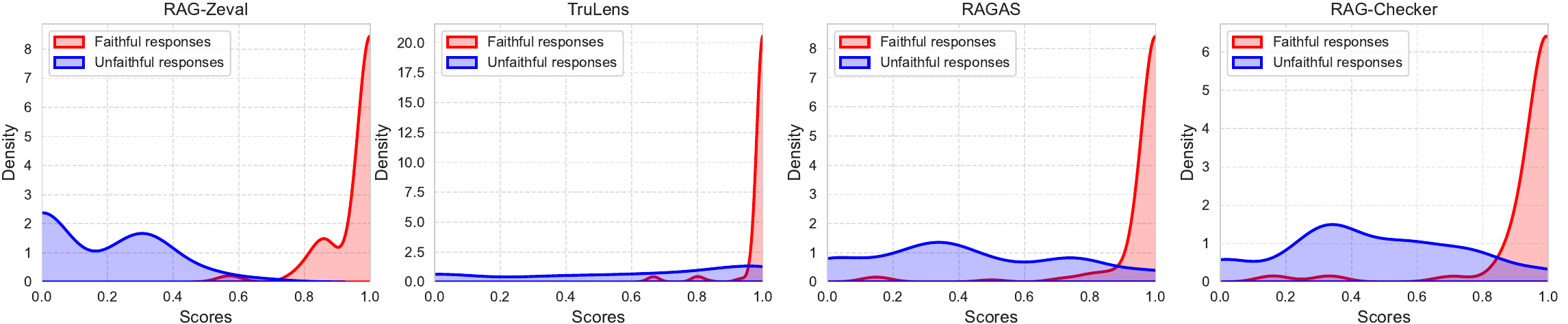}
\caption{The density distribution of the scores assigned by the faithfulness evaluators.The distribution of the faithful and unfaithful responses are marked with red and blue, respectively. TruLens, RAGAS and RAG-Checker are all implemented with \texttt{Qwen2.5-72B-Instruct} as the backbone LLM.} 
\label{fig: score_dist}
\end{figure*}

\begin{table}[t]
\centering
\scalebox{0.7}{
\begin{tabular}{llccc}
\toprule 
\textit{Method} & \textit{Model (-Instruct)} & \textit{Pearson} & \textit{Spearman} & \textit{Kendall} \\ \midrule \midrule
BLEU & -- & 0.302 & 0.305 & 0.236 \\ 
RougL & -- & 0.395 & 0.428 & 0.335 \\ 
BERTScore & -- & 0.350 & 0.437 & 0.341 \\ \midrule
\multirow{3}{*}{ARES} & llama-70b & 0.350 & 0.328 & 0.296 \\
 & qwen-72b & 0.423 & 0.396 & 0.360 \\
 & gpt-4o & 0.382 & 0.370 & 0.333 \\ \midrule
\multirow{3}{*}{TruLens} & llama-70b & 0.428 & 0.453 & 0.366 \\
 & qwen-72b & 0.428 & 0.446 & 0.360 \\
 & gpt-4o & 0.396 & 0.390 & 0.312 \\ \midrule
RAGAS & embedding & 0.411 & 0.432 & 0.283 \\ \midrule 
\multirow{3}{*}{RAG-Checker} & llama-70b & 0.463 & 0.425 & 0.337 \\
 & qwen-72b & 0.495 & 0.465 & 0.375 \\
 & gpt-4o & 0.499 & 0.459 & 0.369 \\ \midrule
SFT & qwen-72b & 0.359 & 0.350 & 0.320 \\ \midrule
\multirow{3}{*}{\begin{tabular}[c]{@{}l@{}}RAG-Zeval \\ w/o RL\end{tabular}} & llama-70b & 0.492 & 0.443 & 0.351 \\
 & qwen-72b & $\textbf{0.521}^{\dagger}$ & $\textbf{0.482}^{\dagger}$ & $\textbf{0.388}^{\dagger}$ \\
 & qwen-7b & 0.427 & 0.367 & 0.312 \\ \midrule
RAG-Zeval & qwen-7b & $\text{0.501}^{\dagger}$ & $\text{0.452}^{\dagger}$ & $\text{0.354}^{\dagger}$ \\ \bottomrule \bottomrule
\end{tabular}}
\caption{Performance on \textbf{correctness} evaluation. Correlation between different methods and human judgments are reported. 
We assess RAGAS \citep{Shahul2023RAGAsAE} with \texttt{Text-Embedding-Ada-002} model \citep{neelakantan2022textcodeembeddingscontrastive} following the original setting. Other settings are the same as Tab.\ref{tab: faithfulness}.
Following \citet{Ru2024}, we only show the metric with the best correlation for each baseline framework. See more details in App. \ref{sec:baseline-appendix}. $\dagger$ indicates the result is statistically significant at the level of 0.01.}
\label{tab: correctness}
\vspace{-0.5cm}
\end{table}

\noindent\textbf{Comparison with Baselines} Table~\ref{tab: faithfulness} and \ref{tab: correctness} present the performances of \texttt{RAG-Zeval} and baseline evaluators. Generally, the claim-based methods outperform non-claim-based ones. For both benchmarks, \texttt{RAG-Zeval} has the strongest correlation with human preference in terms of almost all metrics. Despite its compact architecture (7 billion parameters), \texttt{RAG-Zeval} demonstrates superior performance over most baselines built on large-scale LLMs with 10-100$\times$ more parameters. This result validates the effectiveness of our approach for enhancing evaluation capabilities in compact LLMs. 

For in-depth comparison, Fig.~\ref{fig: score_dist} visualizes the distribution of scores assigned by \texttt{RAG-Zeval} and some baselines that give numerical (instead of categorical) predictions for faithfulness evaluation, where we can see the distribution of faithful and unfaithful responses\footnote{The correctness evaluation benchmark is not used here, due to the fact that human annotators only provide relative assessment (e.g., preference ranking) rather than absolute categorical judgments (correct/incorrect labels).}.
While TruLens has the most concentrated distribution near 1 for faithful responses, its distribution for unfaithful responses disperses evenly across the X-axis, indicating its inability to distinguish the unfaithful responses. For faithful responses, \texttt{RAG-Zeval}, RAG-Checker and RAGAS demonstrate similar distributional shapes, particularly showing comparable peakedness near 1. However, \texttt{RAG-Zeval} shows superior discriminative capacity, maintaining clear separation between faithful and unfaithful response distributions.

\noindent\textbf{Comparison with Ablated Variants} As shown in Tab.~\ref{tab: faithfulness} and \ref{tab: correctness}, SFT on ground-truth ranking exhibits the worst performances among the LLM-based methods, implying the significance of intermediate reasoning. 
On the other hand, even without further training, the non-RL variant of our approach maintains robust evaluation performance. This suggests that our rule-guided generation approach, which generates the reasoning trajectories for evaluation end-to-end, can effectively harness the LLMs' reasoning capabilities to achieve superior evaluation performances. This advantage gets more prominent through RL training, ultimately leading to superior or comparable results compared to large-scale LLM-based baselines. 
More discussion on rule-guided reasoning is in \S~\ref{sec:rule_guided_reasoning}.

\section{Analysis}

The following problems are discussed. 1) How reinforcement learning stimulates the model's evaluation ability. 2) What is the effect of the task complexity represented by training objective and data. 3) What is the effectiveness of the rule-guided reasoning. Here we employ the correctness benchmark as the testbed, since it is more challenging.

\subsection{Self-Evolution of Evaluation Abilities}
\begin{figure}[ht]
\centering
\includegraphics[width=1\linewidth]{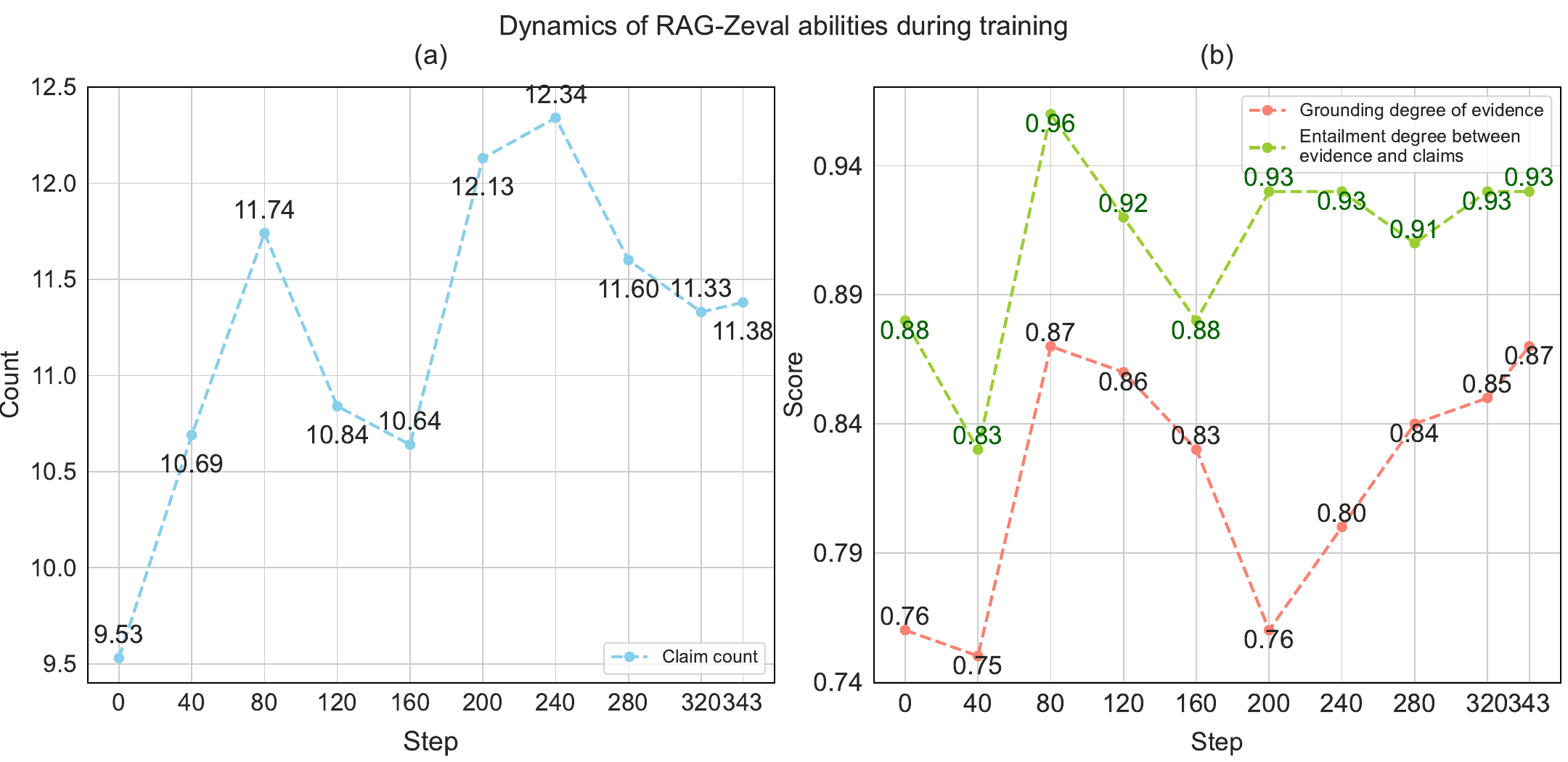}
\caption{(a) shows the changes of decomposed claim count, while (b) presents the evolution of abilities of evidence extraction and supportiveness judgment throughout the RL training process. The statistics are based on the rollout samples during training.}
\label{fig:evolution}
\end{figure}

\begin{table}[t]
\centering
\scalebox{0.65}{
\begin{tabular}{rlll}
\toprule
\multicolumn{1}{c}{Objective} & \textit{Pearson} & \textit{Spearman} & \textit{Kendall}  \\ \midrule \midrule
\multicolumn{1}{r|}{Ranking (Ours)} &  $\text{0.501}^{\dagger}$ & $\text{0.452}^{\dagger}$ & $\text{0.354}^{\dagger}$ \\ \midrule 
\multicolumn{1}{r|}{Predicting the best} & 0.406 & 0.393 & 0.311  \\
 \bottomrule
\end{tabular}
}
\caption{
Comparison of two training objectives. The results are obtained by averaging across 5 runs.
}
\label{tab:objective_comparison}
\vspace{-0.1cm}
\end{table}

\begin{table}[t]
\centering
\scalebox{0.65}{
\begin{tabular}{rlll}
\toprule
\multicolumn{1}{c}{Data configuration} & \textit{Pearson} & \textit{Spearman} & \textit{Kendall}  \\ \midrule \midrule
\multicolumn{1}{r|}{\makecell{\quad\quad\quad Curriculum learning \\ (first 3 and then 4 responses)}} &  $\text{0.501}^{\dagger}$ & $\text{0.452}^{\dagger}$ & $\text{0.354}^{\dagger}$ \\ \midrule 
\multicolumn{1}{r|}{Static (3 responses)} &  0.457 & 0.433 & 0.339 \\ \midrule 
\multicolumn{1}{r|}{Static (4 responses)} & 0.450 & 0.402 & 0.314  \\
 \bottomrule
\end{tabular}
}
\caption{
Comparison of different data configuration. The results are obtained by averaging across 5 runs.
}
\label{tab:data_comparison}
\vspace{-0.3cm}
\end{table}

\begin{table*}[ht]
\centering
\scalebox{0.7}{
\begin{tabular}{rllll}
\toprule
\multicolumn{1}{r}{Method} & Generation & \textit{Pearson} & \textit{Spearman} & \textit{Kendall}  \\ \midrule \midrule
\multicolumn{1}{r|}{Ours} &  The complete reasoning trajectory specified in Fig.\ref{fig:prompt}&  $\text{0.501}^{\dagger}$ & $\text{0.452}^{\dagger}$ & $\text{0.354}^{\dagger}$ \\ 
\multicolumn{1}{r|}{w/o evidence} &  The reasoning trajectory without evidence and analysis &  0.489 & 0.452 & 0.353\\ \midrule 
\multicolumn{1}{r|}{SFT} &  Response ranking result & 0.359 & 0.350 & 0.320  \\
 \bottomrule
\end{tabular}
}
\caption{
Results of methods with different evaluation pattern,  obtained by averaging across 5 runs.
}
\label{tab:reasoning_comparison}
\vspace{-0.5cm}
\end{table*}

To investigate how the model abilities evolve over the RL training process, we continuously monitor the model's behaviors in claim decomposition, evidence extraction, supportiveness judgment. Their dynamics is plotted in Fig.~\ref{fig:evolution}.

The blue line depicts the average number of claims decomposed from a response, which initially exhibits a sharp increase and ends at a stable level. As finer-grained claim decomposition enables more discriminative comparisons among candidate responses, the model learns increasingly comprehensive claim decomposition for enhanced ranking performance. As evidenced by the case study in Fig.~\ref{fig:case_study_norl} and \ref{fig:case_study_rl}, the already trained checkpoint (at step 343) provides a more comprehensive decomposition, whereas some claims by the untrained checkpoint (at step 0) amalgamate atomic claims that should be addressed separately.

For each extracted evidence, we measure its grounding degree in terms of the normalized length of its longest common substring (similar to evidence reward definition in \S\ref{sec:rule_based_rl}). For each supported claim generated by the model, we quantify the degree of textual entailment between its corresponding evidence and the claim itself, using AlignScore\footnote{The evidence is input as \texttt{context}, and the claims is as \texttt{claim}. \texttt{context} and \texttt{claim} are two arguments for AlignScore which measures how likely \texttt{context} would entails \texttt{claim}.}\cite{zha-etal-2023-alignscore}. The grounding (red line) and entailment (green line) degrees both experience a notable growth, implying that the model's capabilities of evidence extraction and supportiveness judgment get improved through the RL training.

Overall, \textit{reinforcement learning effectively incentivizes the development of reasoning capabilities essential for responses evaluation}, consequently improving final evaluation performance.

\noindent\textbf{}
\subsection{Effect of Task Complexity}
\label{sec:ranking_effectiveness}
\noindent\textbf{Ranking-based Objective} 
We simplify the ranking-based accuracy reward (Eq.~\ref{eq:acc_reward}) as
\begin{equation}
\resizebox{0.9\linewidth}{!}{%
    $r_a = \left\{
    \begin{array}{ll}
    1,  \,\text{if} &S(\bm{y}_{\alpha_i}) > S(\bm{y}_{\alpha_j}), \alpha_i = max\{\alpha\},
    \\&\forall \bm{y}_{\alpha_j} \in \{\bm{y}\}\, \text{and}\,\, \bm{y}_{\alpha_i} \neq \bm{y}_{\alpha_j}\\
    0,  \,&\text{otherwise}
    \end{array}
    \right.$}
\label{eq:simplified_reward}
\end{equation}
indicating that an accuracy reward of $\mathbf{1}$ is earned if the model assigns the highest evaluation score to $\bm{y}$ with the largest $\alpha$ value. This simplified formulation, similar to the one used by \citet{Liu2025InferenceTimeSF}, does not require an correct ranking over the entire set of candidate responses. The comparison between the two formulation is shown in Tab.~\ref{tab:objective_comparison}. The model trained with the simplified accuracy reward suffers a notable performance drop. This implies that \textit{reducing task complexity may diminish the incentive for the model to develop enhanced evaluation capabilities}. Because a finer-grained and more discriminative assessment of responses is crucial for achieving a comprehensive ranking.

\noindent\textbf{Curriculum Learning}
During the RL training, we organize the training data in a way that the complexity of the ranking task escalates as the training advances. To study the effect of this practice, we train models with the following two static data organization---the training instance across all epochs consistently contains 3 or 4 candidate responses for ranking.
As illustrated in Tab.~\ref{tab:data_comparison}, the curriculum learning-based configuration has the best performance. Its improvement over the static one with 3 responses further corroborates above finding that increased task complexity may help ability acquisition. However, the static configurations with 4 response performs worst. We found it earns a much lower average combined reward than the curriculum learning-based configuration in the first epoch (seen in App.\ref{sec:training_dynamics}). Employing overly challenging task objective in the initial training stage may suppress model learning, as it is less possible to find a valid rollout and the model then hardly receives positive feedback during the training.

\subsection{Effectiveness of Rule-Guided Reasoning}
\label{sec:rule_guided_reasoning}

Results in \S~\ref{sec:main_experiment} demonstrates that \texttt{RAG-Zeval} outperforms direct \texttt{SFT} on ground-truth ranking of responses. To better illustrate the significance of our rule-guided reasoning, we further introduce an intermediate variant between the above two methods--remove the requirement to provide supporting evidence and corresponding analysis, while maintaining all other settings consistent with \texttt{RAG-Zeval}.

As shown in Tab.~\ref{tab:reasoning_comparison}, their performances are positively correlated with the level of detail of their generation, which substantiates the advantage of the rule-guided reasoning. Analogues to Chain-of-Thought \cite{10.5555/3600270.3602070}, \texttt{RAG-Zeval} benefits from the stepwise reasoning in its evaluation trajectories. Also, detailed evaluation processes offer better explainability behind the model predictions.

\section{Conclusion}
In this work, we introduce \texttt{RAG-Zeval}, a novel rule-guided evaluation framework that enables compact LLMs to perform end-to-end, interpretable assessment of RAG system outputs. Our approach trains evaluators via reinforcement learning with a novel ranking-based objective, bypassing the requirement for human-annotated data. Through comprehensive experiments on benchmarks of faithfulness and correctness evaluation, we demonstrate that our approach achieves strong alignment with human judgments, outperforming current large-scale LLM-based baselines while maintaining a much smaller model scale. 
The result highlights the potential of compact, reasoning-driven evaluators for scalable and transparent RAG evaluation.
\section*{Limitations}
This work has several limitations that point to avenues for future improvement. Although our approach employs smaller models compared to direct use of large LLMs, the RL training process still requires considerable computational resources and access to high-performance hardware, which may not be available to all researchers. In addition, our current experiments are primarily conducted in English and on general-domain datasets; the generalizability of the evaluator to other languages or specific domains remains to be explored. Further validation on multilingual and domain-specific benchmarks would strengthen the robustness and applicability of our method.

Additionally, our experiments run on static datasets, which may not capture real-world dynamic interactions well (e.g., adversarial inputs, evolving user preferences). Further investigation of its performance in real-world environments is essential prior to deployment, to ensure unbiased and accurate judgments.

\section*{Ethical Considerations}
Our RL-based RAG evaluation framework also raises several ethical considerations. The computational requirements, though reduced compared to large LLMs, may still create barriers for less-resourced groups, potentially exacerbating inequities in access to advanced evaluation tools. Moreover, automated evaluation should not be viewed as a substitute for human oversight, especially in high-stakes or sensitive applications, as it may overlook nuanced ethical or contextual factors. Besides, if the synthetic or training data used for evaluator construction contains biases or unrepresentative patterns, these biases may be propagated in the evaluation results. Responsible deployment requires ongoing attention to these issues and a commitment to transparency and fairness


\bibliography{custom}

\begin{thebibliography}{44}
\providecommand{\natexlab}[1]{#1}

\bibitem[{Asai et~al.(2024)Asai, Wu, Wang, Sil, and Hajishirzi}]{asai2024selfrag}
Akari Asai, Zeqiu Wu, Yizhong Wang, Avirup Sil, and Hannaneh Hajishirzi. 2024.
\newblock \href {https://openreview.net/forum?id=hSyW5go0v8} {Self-{RAG}: Learning to retrieve, generate, and critique through self-reflection}.
\newblock In \emph{The Twelfth International Conference on Learning Representations}.

\bibitem[{Bengio et~al.(2009)Bengio, Louradour, Collobert, and Weston}]{10.1145/1553374.1553380}
Yoshua Bengio, J\'{e}r\^{o}me Louradour, Ronan Collobert, and Jason Weston. 2009.
\newblock \href {https://doi.org/10.1145/1553374.1553380} {Curriculum learning}.
\newblock In \emph{Proceedings of the 26th Annual International Conference on Machine Learning}, ICML '09, page 41–48, New York, NY, USA. Association for Computing Machinery.

\bibitem[{DeepSeek-AI et~al.(2025)DeepSeek-AI, Guo, Yang, Zhang, Song, Zhang, Xu, Zhu, Ma, Wang, Bi, Zhang, Yu, Wu, Wu, Gou, Shao, Li, Gao, Liu, Xue, Wang, Wu, Feng, Lu, Zhao, Deng, Zhang, Ruan, Dai, Chen, Ji, Li, Lin, Dai, Luo, Hao, Chen, Li, Zhang, Bao, Xu, Wang, Ding, Xin, Gao, Qu, Li, Guo, Li, Wang, Chen, Yuan, Qiu, Li, Cai, Ni, Liang, Chen, Dong, Hu, Gao, Guan, Huang, Yu, Wang, Zhang, Zhao, Wang, Zhang, Xu, Xia, Zhang, Zhang, Tang, Li, Wang, Li, Tian, Huang, Zhang, Wang, Chen, Du, Ge, Zhang, Pan, Wang, Chen, Jin, Chen, Lu, Zhou, Chen, Ye, Wang, Yu, Zhou, Pan, Li, Zhou, Wu, Ye, Yun, Pei, Sun, Wang, Zeng, Zhao, Liu, Liang, Gao, Yu, Zhang, Xiao, An, Liu, Wang, Chen, Nie, Cheng, Liu, Xie, Liu, Yang, Li, Su, Lin, Li, Jin, Shen, Chen, Sun, Wang, Song, Zhou, Wang, Shan, Li, Wang, Wei, Zhang, Xu, Li, Zhao, Sun, Wang, Yu, Zhang, Shi, Xiong, He, Piao, Wang, Tan, Ma, Liu, Guo, Ou, Wang, Gong, Zou, He, Xiong, Luo, You, Liu, Zhou, Zhu, Xu, Huang, Li, Zheng, Zhu, Ma, Tang, Zha, Yan, Ren, Ren, Sha, Fu, Xu, Xie, Zhang,
  Hao, Ma, Yan, Wu, Gu, Zhu, Liu, Li, Xie, Song, Pan, Huang, Xu, Zhang, and Zhang}]{deepseekai2025deepseekr1incentivizingreasoningcapability}
DeepSeek-AI, Daya Guo, Dejian Yang, Haowei Zhang, Junxiao Song, Ruoyu Zhang, Runxin Xu, Qihao Zhu, Shirong Ma, Peiyi Wang, Xiao Bi, Xiaokang Zhang, Xingkai Yu, Yu~Wu, Z.~F. Wu, Zhibin Gou, Zhihong Shao, Zhuoshu Li, Ziyi Gao, and 181 others. 2025.
\newblock \href {https://arxiv.org/abs/2501.12948} {Deepseek-r1: Incentivizing reasoning capability in llms via reinforcement learning}.
\newblock \emph{Preprint}, arXiv:2501.12948.

\bibitem[{Fan et~al.(2024)Fan, Ding, Ning, Wang, Li, Yin, Chua, and Li}]{fan2024surveyragmeetingllms}
Wenqi Fan, Yujuan Ding, Liangbo Ning, Shijie Wang, Hengyun Li, Dawei Yin, Tat-Seng Chua, and Qing Li. 2024.
\newblock \href {https://arxiv.org/abs/2405.06211} {A survey on rag meeting llms: Towards retrieval-augmented large language models}.
\newblock \emph{Preprint}, arXiv:2405.06211.

\bibitem[{Ferrara et~al.(2024)Ferrara, Ethan-Tonic, and Ozturk}]{ferrara2024ragtriad}
J.~Ferrara, Ethan-Tonic, and O.~M. Ozturk. 2024.
\newblock The rag triad.
\newblock \url{https://www.trulens.org/trulens_eval/core_concepts_rag_triad/}.

\bibitem[{Gao et~al.(2024)Gao, Xiong, Gao, Jia, Pan, Bi, Dai, Sun, Wang, and Wang}]{gao2024retrievalaugmentedgenerationlargelanguage}
Yunfan Gao, Yun Xiong, Xinyu Gao, Kangxiang Jia, Jinliu Pan, Yuxi Bi, Yi~Dai, Jiawei Sun, Meng Wang, and Haofen Wang. 2024.
\newblock \href {https://arxiv.org/abs/2312.10997} {Retrieval-augmented generation for large language models: A survey}.
\newblock \emph{Preprint}, arXiv:2312.10997.

\bibitem[{Grattafiori et~al.(2024)Grattafiori, Dubey, Jauhri, Pandey, Kadian, Al-Dahle, Letman, Mathur, Schelten, Vaughan, Yang, Fan, Goyal, Hartshorn, Yang, Mitra, Sravankumar, Korenev, Hinsvark, Rao, Zhang, Rodriguez, Gregerson, Spataru, Roziere, Biron, Tang, Chern, Caucheteux, Nayak, Bi, Marra, McConnell, Keller, Touret, Wu, Wong, Ferrer, Nikolaidis, Allonsius, Song, Pintz, Livshits, Wyatt, Esiobu, Choudhary, Mahajan, Garcia-Olano, Perino, Hupkes, Lakomkin, AlBadawy, Lobanova, Dinan, Smith, Radenovic, Guzmán, Zhang, Synnaeve, Lee, Anderson, Thattai, Nail, Mialon, Pang, Cucurell, Nguyen, Korevaar, Xu, Touvron, Zarov, Ibarra, Kloumann, Misra, Evtimov, Zhang, Copet, Lee, Geffert, Vranes, Park, Mahadeokar, Shah, van~der Linde, Billock, Hong, Lee, Fu, Chi, Huang, Liu, Wang, Yu, Bitton, Spisak, Park, Rocca, Johnstun, Saxe, Jia, Alwala, Prasad, Upasani, Plawiak, Li, Heafield, Stone, El-Arini, Iyer, Malik, Chiu, Bhalla, Lakhotia, Rantala-Yeary, van~der Maaten, Chen, Tan, Jenkins, Martin, Madaan, Malo, Blecher,
  Landzaat, de~Oliveira, Muzzi, Pasupuleti, Singh, Paluri, Kardas, Tsimpoukelli, Oldham, Rita, Pavlova, Kambadur, Lewis, Si, Singh, Hassan, Goyal, Torabi, Bashlykov, Bogoychev, Chatterji, Zhang, Duchenne, Çelebi, Alrassy, Zhang, Li, Vasic, Weng, Bhargava, Dubal, Krishnan, Koura, Xu, He, Dong, Srinivasan, Ganapathy, Calderer, Cabral, Stojnic, Raileanu, Maheswari, Girdhar, Patel, Sauvestre, Polidoro, Sumbaly, Taylor, Silva, Hou, Wang, Hosseini, Chennabasappa, Singh, Bell, Kim, Edunov, Nie, Narang, Raparthy, Shen, Wan, Bhosale, Zhang, Vandenhende, Batra, Whitman, Sootla, Collot, Gururangan, Borodinsky, Herman, Fowler, Sheasha, Georgiou, Scialom, Speckbacher, Mihaylov, Xiao, Karn, Goswami, Gupta, Ramanathan, Kerkez, Gonguet, Do, Vogeti, Albiero, Petrovic, Chu, Xiong, Fu, Meers, Martinet, Wang, Wang, Tan, Xia, Xie, Jia, Wang, Goldschlag, Gaur, Babaei, Wen, Song, Zhang, Li, Mao, Coudert, Yan, Chen, Papakipos, Singh, Srivastava, Jain, Kelsey, Shajnfeld, Gangidi, Victoria, Goldstand, Menon, Sharma, Boesenberg,
  Baevski, Feinstein, Kallet, Sangani, Teo, Yunus, Lupu, Alvarado, Caples, Gu, Ho, Poulton, Ryan, Ramchandani, Dong, Franco, Goyal, Saraf, Chowdhury, Gabriel, Bharambe, Eisenman, Yazdan, James, Maurer, Leonhardi, Huang, Loyd, Paola, Paranjape, Liu, Wu, Ni, Hancock, Wasti, Spence, Stojkovic, Gamido, Montalvo, Parker, Burton, Mejia, Liu, Wang, Kim, Zhou, Hu, Chu, Cai, Tindal, Feichtenhofer, Gao, Civin, Beaty, Kreymer, Li, Adkins, Xu, Testuggine, David, Parikh, Liskovich, Foss, Wang, Le, Holland, Dowling, Jamil, Montgomery, Presani, Hahn, Wood, Le, Brinkman, Arcaute, Dunbar, Smothers, Sun, Kreuk, Tian, Kokkinos, Ozgenel, Caggioni, Kanayet, Seide, Florez, Schwarz, Badeer, Swee, Halpern, Herman, Sizov, Guangyi, Zhang, Lakshminarayanan, Inan, Shojanazeri, Zou, Wang, Zha, Habeeb, Rudolph, Suk, Aspegren, Goldman, Zhan, Damlaj, Molybog, Tufanov, Leontiadis, Veliche, Gat, Weissman, Geboski, Kohli, Lam, Asher, Gaya, Marcus, Tang, Chan, Zhen, Reizenstein, Teboul, Zhong, Jin, Yang, Cummings, Carvill, Shepard, McPhie,
  Torres, Ginsburg, Wang, Wu, U, Saxena, Khandelwal, Zand, Matosich, Veeraraghavan, Michelena, Li, Jagadeesh, Huang, Chawla, Huang, Chen, Garg, A, Silva, Bell, Zhang, Guo, Yu, Moshkovich, Wehrstedt, Khabsa, Avalani, Bhatt, Mankus, Hasson, Lennie, Reso, Groshev, Naumov, Lathi, Keneally, Liu, Seltzer, Valko, Restrepo, Patel, Vyatskov, Samvelyan, Clark, Macey, Wang, Hermoso, Metanat, Rastegari, Bansal, Santhanam, Parks, White, Bawa, Singhal, Egebo, Usunier, Mehta, Laptev, Dong, Cheng, Chernoguz, Hart, Salpekar, Kalinli, Kent, Parekh, Saab, Balaji, Rittner, Bontrager, Roux, Dollar, Zvyagina, Ratanchandani, Yuvraj, Liang, Alao, Rodriguez, Ayub, Murthy, Nayani, Mitra, Parthasarathy, Li, Hogan, Battey, Wang, Howes, Rinott, Mehta, Siby, Bondu, Datta, Chugh, Hunt, Dhillon, Sidorov, Pan, Mahajan, Verma, Yamamoto, Ramaswamy, Lindsay, Lindsay, Feng, Lin, Zha, Patil, Shankar, Zhang, Zhang, Wang, Agarwal, Sajuyigbe, Chintala, Max, Chen, Kehoe, Satterfield, Govindaprasad, Gupta, Deng, Cho, Virk, Subramanian, Choudhury,
  Goldman, Remez, Glaser, Best, Koehler, Robinson, Li, Zhang, Matthews, Chou, Shaked, Vontimitta, Ajayi, Montanez, Mohan, Kumar, Mangla, Ionescu, Poenaru, Mihailescu, Ivanov, Li, Wang, Jiang, Bouaziz, Constable, Tang, Wu, Wang, Wu, Gao, Kleinman, Chen, Hu, Jia, Qi, Li, Zhang, Zhang, Adi, Nam, Yu, Wang, Zhao, Hao, Qian, Li, He, Rait, DeVito, Rosnbrick, Wen, Yang, Zhao, and Ma}]{grattafiori2024llama3herdmodels}
Aaron Grattafiori, Abhimanyu Dubey, Abhinav Jauhri, Abhinav Pandey, Abhishek Kadian, Ahmad Al-Dahle, Aiesha Letman, Akhil Mathur, Alan Schelten, Alex Vaughan, Amy Yang, Angela Fan, Anirudh Goyal, Anthony Hartshorn, Aobo Yang, Archi Mitra, Archie Sravankumar, Artem Korenev, Arthur Hinsvark, and 542 others. 2024.
\newblock \href {https://arxiv.org/abs/2407.21783} {The llama 3 herd of models}.
\newblock \emph{Preprint}, arXiv:2407.21783.

\bibitem[{Gu et~al.(2024)Gu, Jiang, Shi, Tan, Zhai, Xu, Li, Shen, Ma, Liu, Wang, and Guo}]{Gu2024ASO}
Jiawei Gu, Xuhui Jiang, Zhichao Shi, Hexiang Tan, Xuehao Zhai, Chengjin Xu, Wei Li, Yinghan Shen, Shengjie Ma, Honghao Liu, Yuanzhuo Wang, and Jian Guo. 2024.
\newblock \href {https://api.semanticscholar.org/CorpusID:274234014} {A survey on llm-as-a-judge}.
\newblock \emph{ArXiv}, abs/2411.15594.

\bibitem[{Guan et~al.(2025)Guan, Zhang, Liu, Shang, Sun, Zhu, Yang, and Yang}]{guan2025rstarmathsmallllmsmaster}
Xinyu Guan, Li~Lyna Zhang, Yifei Liu, Ning Shang, Youran Sun, Yi~Zhu, Fan Yang, and Mao Yang. 2025.
\newblock \href {https://arxiv.org/abs/2501.04519} {rstar-math: Small llms can master math reasoning with self-evolved deep thinking}.
\newblock \emph{Preprint}, arXiv:2501.04519.

\bibitem[{Holtzman et~al.(2019)Holtzman, Buys, Du, Forbes, and Choi}]{Holtzman2019TheCC}
Ari Holtzman, Jan Buys, Li~Du, Maxwell Forbes, and Yejin Choi. 2019.
\newblock \href {https://api.semanticscholar.org/CorpusID:127986954} {The curious case of neural text degeneration}.
\newblock \emph{ArXiv}, abs/1904.09751.

\bibitem[{Ispas et~al.(2025)Ispas, Simon, Caspani, and Guigue}]{Ispas2025TowardsLA}
Alex-Răzvan Ispas, Charles-Elie Simon, Fabien Caspani, and Vincent Guigue. 2025.
\newblock \href {https://api.semanticscholar.org/CorpusID:277150863} {Towards lighter and robust evaluation for retrieval augmented generation}.
\newblock \emph{The Next Frontier in Reliable AI": Workshop on ICLR 2025}, abs/2503.16161.

\bibitem[{Kwiatkowski et~al.(2019)Kwiatkowski, Palomaki, Redfield, Collins, Parikh, Alberti, Epstein, Polosukhin, Devlin, Lee, Toutanova, Jones, Kelcey, Chang, Dai, Uszkoreit, Le, and Petrov}]{Kwiatkowski2019NaturalQA}
Tom Kwiatkowski, Jennimaria Palomaki, Olivia Redfield, Michael Collins, Ankur~P. Parikh, Chris Alberti, Danielle Epstein, Illia Polosukhin, Jacob Devlin, Kenton Lee, Kristina Toutanova, Llion Jones, Matthew Kelcey, Ming-Wei Chang, Andrew~M. Dai, Jakob Uszkoreit, Quoc~V. Le, and Slav Petrov. 2019.
\newblock \href {https://api.semanticscholar.org/CorpusID:86611921} {Natural questions: A benchmark for question answering research}.
\newblock \emph{Transactions of the Association for Computational Linguistics}, 7:453--466.

\bibitem[{Lewis et~al.(2021)Lewis, Perez, Piktus, Petroni, Karpukhin, Goyal, Küttler, Lewis, tau Yih, Rocktäschel, Riedel, and Kiela}]{lewis2021retrievalaugmentedgenerationknowledgeintensivenlp}
Patrick Lewis, Ethan Perez, Aleksandra Piktus, Fabio Petroni, Vladimir Karpukhin, Naman Goyal, Heinrich Küttler, Mike Lewis, Wen tau Yih, Tim Rocktäschel, Sebastian Riedel, and Douwe Kiela. 2021.
\newblock \href {https://arxiv.org/abs/2005.11401} {Retrieval-augmented generation for knowledge-intensive nlp tasks}.
\newblock \emph{Preprint}, arXiv:2005.11401.

\bibitem[{Li et~al.(2025)Li, Zhang, Li, Luo, Moustafa, Wu, Glass, and Meng}]{li2025generatediscriminateevolveenhancing}
Kun Li, Tianhua Zhang, Yunxiang Li, Hongyin Luo, Abdalla Moustafa, Xixin Wu, James Glass, and Helen Meng. 2025.
\newblock \href {https://arxiv.org/abs/2503.01695} {Generate, discriminate, evolve: Enhancing context faithfulness via fine-grained sentence-level self-evolution}.
\newblock \emph{Preprint}, arXiv:2503.01695.

\bibitem[{Li et~al.(2024)Li, Zhang, Wu, Luo, Glass, and Meng}]{li2024decodinggraphsfaithfulsound}
Kun Li, Tianhua Zhang, Xixin Wu, Hongyin Luo, James Glass, and Helen Meng. 2024.
\newblock \href {https://arxiv.org/abs/2410.18415} {Decoding on graphs: Faithful and sound reasoning on knowledge graphs through generation of well-formed chains}.
\newblock \emph{Preprint}, arXiv:2410.18415.

\bibitem[{Li et~al.(2022)Li, Holtzman, Fried, Liang, Eisner, Hashimoto, Zettlemoyer, and Lewis}]{Li2022ContrastiveDO}
Xiang~Lisa Li, Ari Holtzman, Daniel Fried, Percy Liang, Jason Eisner, Tatsunori Hashimoto, Luke Zettlemoyer, and Mike Lewis. 2022.
\newblock \href {https://aclanthology.org/2023.acl-long.687/} {Contrastive decoding: Open-ended text generation as optimization}.
\newblock In \emph{Annual Meeting of the Association for Computational Linguistics}.

\bibitem[{Lightman et~al.(2024)Lightman, Kosaraju, Burda, Edwards, Baker, Lee, Leike, Schulman, Sutskever, and Cobbe}]{lightman2024lets}
Hunter Lightman, Vineet Kosaraju, Yuri Burda, Harrison Edwards, Bowen Baker, Teddy Lee, Jan Leike, John Schulman, Ilya Sutskever, and Karl Cobbe. 2024.
\newblock \href {https://openreview.net/forum?id=v8L0pN6EOi} {Let's verify step by step}.
\newblock In \emph{The Twelfth International Conference on Learning Representations}.

\bibitem[{Lin(2004)}]{lin-2004-rouge}
Chin-Yew Lin. 2004.
\newblock \href {https://aclanthology.org/W04-1013/} {{ROUGE}: A package for automatic evaluation of summaries}.
\newblock In \emph{Text Summarization Branches Out}, pages 74--81, Barcelona, Spain. Association for Computational Linguistics.

\bibitem[{Liu et~al.(2025{\natexlab{a}})Liu, Chen, Li, Qi, Pang, Du, Lee, and Lin}]{liu2025understandingr1zeroliketrainingcritical}
Zichen Liu, Changyu Chen, Wenjun Li, Penghui Qi, Tianyu Pang, Chao Du, Wee~Sun Lee, and Min Lin. 2025{\natexlab{a}}.
\newblock \href {https://arxiv.org/abs/2503.20783} {Understanding r1-zero-like training: A critical perspective}.
\newblock \emph{Preprint}, arXiv:2503.20783.

\bibitem[{Liu et~al.(2025{\natexlab{b}})Liu, Wang, Xu, Ma, Ruan, Li, Liu, and Wu}]{liu2025inferencetimescalinggeneralistreward}
Zijun Liu, Peiyi Wang, Runxin Xu, Shirong Ma, Chong Ruan, Peng Li, Yang Liu, and Yu~Wu. 2025{\natexlab{b}}.
\newblock \href {https://arxiv.org/abs/2504.02495} {Inference-time scaling for generalist reward modeling}.
\newblock \emph{Preprint}, arXiv:2504.02495.

\bibitem[{Liu et~al.(2025{\natexlab{c}})Liu, Wang, Xu, Ma, Ruan, Li, Liu, and Wu}]{Liu2025InferenceTimeSF}
Zijun Liu, Peiyi Wang, Runxin Xu, Shirong Ma, Chong Ruan, Peng Li, Yang Liu, and Yu~Wu. 2025{\natexlab{c}}.
\newblock \href {https://api.semanticscholar.org/CorpusID:277510339} {Inference-time scaling for generalist reward modeling}.

\bibitem[{Loshchilov and Hutter(2017)}]{Loshchilov2017DecoupledWD}
Ilya Loshchilov and Frank Hutter. 2017.
\newblock \href {https://api.semanticscholar.org/CorpusID:53592270} {Decoupled weight decay regularization}.
\newblock In \emph{International Conference on Learning Representations}.

\bibitem[{Manakul et~al.(2023)Manakul, Liusie, and Gales}]{manakul-etal-2023-selfcheckgpt}
Potsawee Manakul, Adian Liusie, and Mark Gales. 2023.
\newblock \href {https://doi.org/10.18653/v1/2023.emnlp-main.557} {{S}elf{C}heck{GPT}: Zero-resource black-box hallucination detection for generative large language models}.
\newblock In \emph{Proceedings of the 2023 Conference on Empirical Methods in Natural Language Processing}, pages 9004--9017, Singapore. Association for Computational Linguistics.

\bibitem[{Narvekar et~al.(2020)Narvekar, Peng, Leonetti, Sinapov, Taylor, and Stone}]{10.5555/3455716.3455897}
Sanmit Narvekar, Bei Peng, Matteo Leonetti, Jivko Sinapov, Matthew~E. Taylor, and Peter Stone. 2020.
\newblock Curriculum learning for reinforcement learning domains: a framework and survey.
\newblock 21(1).

\bibitem[{Neelakantan et~al.(2022)Neelakantan, Xu, Puri, Radford, Han, Tworek, Yuan, Tezak, Kim, Hallacy, Heidecke, Shyam, Power, Nekoul, Sastry, Krueger, Schnurr, Such, Hsu, Thompson, Khan, Sherbakov, Jang, Welinder, and Weng}]{neelakantan2022textcodeembeddingscontrastive}
Arvind Neelakantan, Tao Xu, Raul Puri, Alec Radford, Jesse~Michael Han, Jerry Tworek, Qiming Yuan, Nikolas Tezak, Jong~Wook Kim, Chris Hallacy, Johannes Heidecke, Pranav Shyam, Boris Power, Tyna~Eloundou Nekoul, Girish Sastry, Gretchen Krueger, David Schnurr, Felipe~Petroski Such, Kenny Hsu, and 6 others. 2022.
\newblock \href {https://arxiv.org/abs/2201.10005} {Text and code embeddings by contrastive pre-training}.
\newblock \emph{Preprint}, arXiv:2201.10005.

\bibitem[{OpenAI et~al.(2024)OpenAI, Achiam, Adler, Agarwal, Ahmad, Akkaya, Aleman, Almeida, Altenschmidt, Altman, Anadkat, Avila, Babuschkin, Balaji, Balcom, Baltescu, Bao, Bavarian, Belgum, Bello, Berdine, Bernadett-Shapiro, Berner, Bogdonoff, Boiko, Boyd, Brakman, Brockman, Brooks, Brundage, Button, Cai, Campbell, Cann, Carey, Carlson, Carmichael, Chan, Chang, Chantzis, Chen, Chen, Chen, Chen, Chen, Chess, Cho, Chu, Chung, Cummings, Currier, Dai, Decareaux, Degry, Deutsch, Deville, Dhar, Dohan, Dowling, Dunning, Ecoffet, Eleti, Eloundou, Farhi, Fedus, Felix, Fishman, Forte, Fulford, Gao, Georges, Gibson, Goel, Gogineni, Goh, Gontijo-Lopes, Gordon, Grafstein, Gray, Greene, Gross, Gu, Guo, Hallacy, Han, Harris, He, Heaton, Heidecke, Hesse, Hickey, Hickey, Hoeschele, Houghton, Hsu, Hu, Hu, Huizinga, Jain, Jain, Jang, Jiang, Jiang, Jin, Jin, Jomoto, Jonn, Jun, Kaftan, Łukasz Kaiser, Kamali, Kanitscheider, Keskar, Khan, Kilpatrick, Kim, Kim, Kim, Kirchner, Kiros, Knight, Kokotajlo, Łukasz Kondraciuk,
  Kondrich, Konstantinidis, Kosic, Krueger, Kuo, Lampe, Lan, Lee, Leike, Leung, Levy, Li, Lim, Lin, Lin, Litwin, Lopez, Lowe, Lue, Makanju, Malfacini, Manning, Markov, Markovski, Martin, Mayer, Mayne, McGrew, McKinney, McLeavey, McMillan, McNeil, Medina, Mehta, Menick, Metz, Mishchenko, Mishkin, Monaco, Morikawa, Mossing, Mu, Murati, Murk, Mély, Nair, Nakano, Nayak, Neelakantan, Ngo, Noh, Ouyang, O'Keefe, Pachocki, Paino, Palermo, Pantuliano, Parascandolo, Parish, Parparita, Passos, Pavlov, Peng, Perelman, de~Avila Belbute~Peres, Petrov, de~Oliveira~Pinto, Michael, Pokorny, Pokrass, Pong, Powell, Power, Power, Proehl, Puri, Radford, Rae, Ramesh, Raymond, Real, Rimbach, Ross, Rotsted, Roussez, Ryder, Saltarelli, Sanders, Santurkar, Sastry, Schmidt, Schnurr, Schulman, Selsam, Sheppard, Sherbakov, Shieh, Shoker, Shyam, Sidor, Sigler, Simens, Sitkin, Slama, Sohl, Sokolowsky, Song, Staudacher, Such, Summers, Sutskever, Tang, Tezak, Thompson, Tillet, Tootoonchian, Tseng, Tuggle, Turley, Tworek, Uribe, Vallone,
  Vijayvergiya, Voss, Wainwright, Wang, Wang, Wang, Ward, Wei, Weinmann, Welihinda, Welinder, Weng, Weng, Wiethoff, Willner, Winter, Wolrich, Wong, Workman, Wu, Wu, Wu, Xiao, Xu, Yoo, Yu, Yuan, Zaremba, Zellers, Zhang, Zhang, Zhao, Zheng, Zhuang, Zhuk, and Zoph}]{openai2024gpt4technicalreport}
OpenAI, Josh Achiam, Steven Adler, Sandhini Agarwal, Lama Ahmad, Ilge Akkaya, Florencia~Leoni Aleman, Diogo Almeida, Janko Altenschmidt, Sam Altman, Shyamal Anadkat, Red Avila, Igor Babuschkin, Suchir Balaji, Valerie Balcom, Paul Baltescu, Haiming Bao, Mohammad Bavarian, Jeff Belgum, and 262 others. 2024.
\newblock \href {https://arxiv.org/abs/2303.08774} {Gpt-4 technical report}.
\newblock \emph{Preprint}, arXiv:2303.08774.

\bibitem[{Papineni et~al.(2002)Papineni, Roukos, Ward, and Zhu}]{papineni-etal-2002-bleu}
Kishore Papineni, Salim Roukos, Todd Ward, and Wei-Jing Zhu. 2002.
\newblock \href {https://doi.org/10.3115/1073083.1073135} {{B}leu: a method for automatic evaluation of machine translation}.
\newblock In \emph{Proceedings of the 40th Annual Meeting of the Association for Computational Linguistics}, pages 311--318, Philadelphia, Pennsylvania, USA. Association for Computational Linguistics.

\bibitem[{Pipitone and Alami(2024)}]{Pipitone2024LegalBenchRAGAB}
Nicholas Pipitone and Ghita~Houir Alami. 2024.
\newblock \href {https://api.semanticscholar.org/CorpusID:271909426} {Legalbench-rag: A benchmark for retrieval-augmented generation in the legal domain}.
\newblock \emph{ArXiv}, abs/2408.10343.

\bibitem[{Qwen(2024)}]{qwen2.5}
Qwen. 2024.
\newblock \href {https://qwenlm.github.io/blog/qwen2.5/} {Qwen2.5: A party of foundation models}.

\bibitem[{Ru et~al.(2024)Ru, Qiu, Hu, Zhang, Shi, Chang, Jiayang, Wang, Sun, Li, Zhang, Wang, Jiang, He, Wang, Liu, Zhang, and Zhang}]{Ru2024}
Dongyu Ru, Lin Qiu, Xiangkun Hu, Tianhang Zhang, Peng Shi, Shuaichen Chang, Cheng Jiayang, Cunxiang Wang, Shichao Sun, Huanyu Li, Zizhao Zhang, Binjie Wang, Jiarong Jiang, Tong He, Zhiguo Wang, Pengfei Liu, Yue Zhang, and Zheng Zhang. 2024.
\newblock \href {https://www.amazon.science/publications/ragchecker-a-fine-grained-framework-for-diagnosing-retrieval-augmented-generation} {Ragchecker: A fine-grained framework for diagnosing retrieval-augmented generation}.
\newblock \emph{NeurIPS}.

\bibitem[{Saad-Falcon et~al.(2024)Saad-Falcon, Khattab, Potts, and Zaharia}]{saad-falcon-etal-2024-ares}
Jon Saad-Falcon, Omar Khattab, Christopher Potts, and Matei Zaharia. 2024.
\newblock \href {https://doi.org/10.18653/v1/2024.naacl-long.20} {{ARES}: An automated evaluation framework for retrieval-augmented generation systems}.
\newblock In \emph{Proceedings of the 2024 Conference of the North American Chapter of the Association for Computational Linguistics: Human Language Technologies (Volume 1: Long Papers)}, pages 338--354, Mexico City, Mexico. Association for Computational Linguistics.

\bibitem[{Shahul et~al.(2023)Shahul, James, Anke, and Schockaert}]{Shahul2023RAGAsAE}
ES~Shahul, Jithin James, Luis~Espinosa Anke, and Steven Schockaert. 2023.
\newblock \href {https://api.semanticscholar.org/CorpusID:263152733} {Ragas: Automated evaluation of retrieval augmented generation}.
\newblock In \emph{Conference of the European Chapter of the Association for Computational Linguistics}.

\bibitem[{Shao et~al.(2024)Shao, Wang, Zhu, Xu, Song, Zhang, Li, Wu, and Guo}]{Shao2024DeepSeekMathPT}
Zhihong Shao, Peiyi Wang, Qihao Zhu, Runxin Xu, Jun-Mei Song, Mingchuan Zhang, Y.~K. Li, Yu~Wu, and Daya Guo. 2024.
\newblock \href {https://api.semanticscholar.org/CorpusID:267412607} {Deepseekmath: Pushing the limits of mathematical reasoning in open language models}.
\newblock \emph{ArXiv}, abs/2402.03300.

\bibitem[{Sheng et~al.(2024)Sheng, Zhang, Ye, Wu, Zhang, Zhang, Peng, Lin, and Wu}]{sheng2024hybridflow}
Guangming Sheng, Chi Zhang, Zilingfeng Ye, Xibin Wu, Wang Zhang, Ru~Zhang, Yanghua Peng, Haibin Lin, and Chuan Wu. 2024.
\newblock Hybridflow: A flexible and efficient rlhf framework.
\newblock \emph{arXiv preprint arXiv: 2409.19256}.

\bibitem[{Shi et~al.(2023)Shi, Han, Lewis, Tsvetkov, Zettlemoyer, and Yih}]{Shi2023TrustingYE}
Weijia Shi, Xiaochuang Han, Mike Lewis, Yulia Tsvetkov, Luke Zettlemoyer, and Scott Yih. 2023.
\newblock \href {https://aclanthology.org/2024.naacl-short.69/} {Trusting your evidence: Hallucinate less with context-aware decoding}.
\newblock In \emph{North American Chapter of the Association for Computational Linguistics}.

\bibitem[{Tang and Yang(2024)}]{tang2024multihoprag}
Yixuan Tang and Yi~Yang. 2024.
\newblock \href {https://arxiv.org/abs/2401.15391} {Multihop-rag: Benchmarking retrieval-augmented generation for multi-hop queries}.
\newblock \emph{Preprint}, arXiv:2401.15391.

\bibitem[{Wei et~al.(2022)Wei, Wang, Schuurmans, Bosma, Ichter, Xia, Chi, Le, and Zhou}]{10.5555/3600270.3602070}
Jason Wei, Xuezhi Wang, Dale Schuurmans, Maarten Bosma, Brian Ichter, Fei Xia, Ed~H. Chi, Quoc~V. Le, and Denny Zhou. 2022.
\newblock Chain-of-thought prompting elicits reasoning in large language models.
\newblock In \emph{Proceedings of the 36th International Conference on Neural Information Processing Systems}, NIPS '22, Red Hook, NY, USA. Curran Associates Inc.

\bibitem[{Yu et~al.(2024)Yu, Gan, Zhang, Tong, Liu, and Liu}]{Yu2024EvaluationOR}
Hao Yu, Aoran Gan, Kai Zhang, Shiwei Tong, Qi~Liu, and Zhaofeng Liu. 2024.
\newblock \href {https://api.semanticscholar.org/CorpusID:269758033} {Evaluation of retrieval-augmented generation: A survey}.
\newblock \emph{ArXiv}, abs/2405.07437.

\bibitem[{Yu et~al.(2025{\natexlab{a}})Yu, Gan, Zhang, Tong, Liu, and Liu}]{Yu_2025}
Hao Yu, Aoran Gan, Kai Zhang, Shiwei Tong, Qi~Liu, and Zhaofeng Liu. 2025{\natexlab{a}}.
\newblock \href {https://doi.org/10.1007/978-981-96-1024-2_8} {\emph{Evaluation of Retrieval-Augmented Generation: A Survey}}, page 102–120.
\newblock Springer Nature Singapore.

\bibitem[{Yu et~al.(2025{\natexlab{b}})Yu, Zhang, Zhu, Yuan, Zuo, Yue, Fan, Liu, Liu, Liu, Lin, Lin, Ma, Sheng, Tong, Zhang, Zhang, Zhang, Zhu, Zhu, Chen, Chen, Wang, Yu, Dai, Song, Wei, Zhou, Liu, Ma, Zhang, Yan, Qiao, Wu, and Wang}]{yu2025dapoopensourcellmreinforcement}
Qiying Yu, Zheng Zhang, Ruofei Zhu, Yufeng Yuan, Xiaochen Zuo, Yu~Yue, Tiantian Fan, Gaohong Liu, Lingjun Liu, Xin Liu, Haibin Lin, Zhiqi Lin, Bole Ma, Guangming Sheng, Yuxuan Tong, Chi Zhang, Mofan Zhang, Wang Zhang, Hang Zhu, and 16 others. 2025{\natexlab{b}}.
\newblock \href {https://arxiv.org/abs/2503.14476} {Dapo: An open-source llm reinforcement learning system at scale}.
\newblock \emph{Preprint}, arXiv:2503.14476.

\bibitem[{Zha et~al.(2023)Zha, Yang, Li, and Hu}]{zha-etal-2023-alignscore}
Yuheng Zha, Yichi Yang, Ruichen Li, and Zhiting Hu. 2023.
\newblock \href {https://doi.org/10.18653/v1/2023.acl-long.634} {{A}lign{S}core: Evaluating factual consistency with a unified alignment function}.
\newblock In \emph{Proceedings of the 61st Annual Meeting of the Association for Computational Linguistics (Volume 1: Long Papers)}, pages 11328--11348, Toronto, Canada. Association for Computational Linguistics.

\bibitem[{Zhang et~al.(2020)Zhang, Kishore*, Wu*, Weinberger, and Artzi}]{bert-score}
Tianyi Zhang, Varsha Kishore*, Felix Wu*, Kilian~Q. Weinberger, and Yoav Artzi. 2020.
\newblock \href {https://openreview.net/forum?id=SkeHuCVFDr} {Bertscore: Evaluating text generation with bert}.
\newblock In \emph{International Conference on Learning Representations}.

\bibitem[{Zhao et~al.(2025)Zhao, Liu, Yang, and Miao}]{zhao2025medrag}
Xuejiao Zhao, Siyan Liu, Su-Yin Yang, and Chunyan Miao. 2025.
\newblock Medrag: Enhancing retrieval-augmented generation with knowledge graph-elicited reasoning for healthcare copilot.
\newblock In \emph{Proceedings of the ACM on Web Conference 2025}, pages 4442--4457.

\bibitem[{Zhong et~al.(2025)Zhong, Shen, Li, Gao, Lu, Chen, Zhang, Zhou, Gu, and Zou}]{zhong2025comprehensivesurveyrewardmodels}
Jialun Zhong, Wei Shen, Yanzeng Li, Songyang Gao, Hua Lu, Yicheng Chen, Yang Zhang, Wei Zhou, Jinjie Gu, and Lei Zou. 2025.
\newblock \href {https://arxiv.org/abs/2504.12328} {A comprehensive survey of reward models: Taxonomy, applications, challenges, and future}.
\newblock \emph{Preprint}, arXiv:2504.12328.

\end{thebibliography}

\newpage
\appendix

\section{Implementation Details}
\label{sec:implementation-appendix}
\subsection{Response Synthesis}
\label{sec:synthesis-appendix}
Starting from Natural Question dataset \cite{Kwiatkowski2019NaturalQA}, we first filter out the instances with a passage that has more than 6,000 tokens. We then randomly select 5,500 instances from the remaining instances. For each $\alpha \in \{0, -0.5, -1, -1.4\}$, we synthesize a response according to Eq.~\ref{eq:cad}, using \texttt{Qwen2.5-7B-Instruct} \cite{qwen2.5}. $P_{LLM} (*\mid \bm{q}, \bm{c})$ and $P_{LLM} (*\mid \bm{q})$ are modeled using in-context learning, and the in-context prompts are shown in Tab.\ref{tab:appendix-prompt}. Greedy search is used for sampling tokens.

\subsection{RAG-Zeval}
\label{sec:ours-implementation-appendix}
We utilize VERL \cite{sheng2024hybridflow}, a open-source library, to apply RL training to the models. The training runs on 8 H20 GPUs and takes approximately 20 hours. The hyperparameters for the training are listed in Tab. \ref{tab:hyperpara}.

\begin{table}[h]
\centering
\scalebox{0.75}{
\begin{tabular}{ll}
\hline
Hyperparameters  \\ \hline
Training batch size & 32 \\
Optimizer & AdamW \\
 &  \cite{Loshchilov2017DecoupledWD}\\
Learning rate & 1e-6 \\
Warmup step & 10 \\
Gradient accumulation step & 1 \\
Learning rate scheduler & Linear \\
KL coefficient & 0.015 \\
Rollout temperature & 1 \\ 
Rollout number & 8 \\ 
Rollout maximum length & 8192 \\ 
Total epoch & 2 \\ \hline
\end{tabular}}
\caption{The settings of hyperparameters used in the RL training.}
\label{tab:hyperpara}
\end{table}

\subsection{Faithfulness Metric}
\label{sec:metric-appendix}
When assessing the faithfulness evaluation performance of different methods, we follow \citet{Shahul2023RAGAsAE} to handle possible \textit{ties} with three scenarios: 
\begin{itemize}
\vspace{-0.15cm}
    \item \textbf{Best-Case}: Measures the frequency of evaluators assigning greater or equal faithfulness scores to good answers compared to poor ones. 
    {
     \setlength{\abovedisplayskip}{0pt}
     \setlength{\belowdisplayskip}{0pt}
    \[
        \mathit{best} = \frac{1}{n} \sum_{i=1}^n \mathbb{I}\left[\mathrm{F}(\text{good}_i) \geq \mathrm{F}(\text{poor}_i)\right]
    \]}
    \item \textbf{Worst-Case}: Computes the frequency of strictly greater faithfulness scores assigned to good answers.
    {
     \setlength{\abovedisplayskip}{0pt}
     \setlength{\belowdisplayskip}{0pt}
    \[
        \mathit{worst} = \frac{1}{n} \sum_{i=1}^n \mathbb{I}\left[\mathrm{F}(\text{good}_i) > \mathrm{F}(\text{poor}_i)\right]
    \]}
    \item \textbf{Middle-Case}: Adopts ternary scoring with a partial point of $0.5$ for ties.
    {
     \setlength{\abovedisplayskip}{0pt}
     \setlength{\belowdisplayskip}{0pt}
    \begin{equation*}
    \begin{aligned}
        \mathit{middle} = \frac{1}{n} \sum_{i=1}^n\{\mathbb{I}[\mathrm{F(good_i)} > \mathrm{F(poor_i)}]\\
    + 0.5\cdot \mathbb{I}[\mathrm{F(good_i)} = \mathrm{F(poor_i)}]\}
    \end{aligned}
    \end{equation*}}
\vspace{-0.3cm}
\end{itemize}

\subsection{Baselines}
\label{sec:baseline-appendix}
For \textbf{standard SFT} baseline, we enhance model generalizability across varying number of candidate answers by randomly partitioning the training data (described in \S \ref{sec:rule_based_rl}) into three subsets: pairwise (2 responses), triplet (3 responses) and quadruplet (4 responses) ranking tasks. Each subset contains approximately equal data volume as reported in Tab.\ref{tab:sft-data-statistics}. The model is trained to reproduce the relative ranking of responses based on their faithfulness with respect to the grounding passage.

\begin{table}[h]
\centering
\scalebox{0.8}{
\begin{tabular}{lcccc}
\toprule
\multicolumn{1}{c}{} & Pairwise & Triplet & Quadruplet & Total \\ \midrule
\multicolumn{1}{l|}{Question} & 647 & 970 & 3883 & 5500 \\
\multicolumn{1}{l|}{Instance} & 3877 & 3874 & 3883 & 11634 \\
\bottomrule
\end{tabular}}
\caption{Data statistics for standard supervised fine-tuning. Each original question includes four generated responses in different faithfulness levels, yielding six pairwise, four triplet, and one quadruplet ranking instance per question. The final row reports deduplicated instance counts.}
\label{tab:sft-data-statistics}
\end{table}

For correctness evaluation, not all baseline evaluation framework has a direct correctness metric. For RAG-Checker \cite{Ru2024}, we report the performance using the \textit{precision} metric, which aligns with our definition of correctness. For baselines without a direct correctness metric, we follow the setting in \citet{Ru2024} to report the best correlation among all metrics in Tab.\ref{tab: correctness}. Tab.\ref{tab:baseline-metrics} shows the complete results of Trulens \cite{ferrara2024ragtriad} and ARES \cite{saad-falcon-etal-2024-ares} using \texttt{Qwen2.5-72B-Instruct}.

\begin{table}[h]
\centering
\scalebox{0.8}{
\begin{tabular}{llccc}
\toprule
Method & Model & \multicolumn{1}{l}{Pearson} & \multicolumn{1}{l}{Spearman} & \multicolumn{1}{l}{Kendall} \\ \midrule
\multirow{2}{*}{ARES} & Relevancy & 0.423 & 0.396 & 0.360 \\
 & Faithfulness & 0.372 & 0.356 & 0.320 \\ \midrule
\multirow{2}{*}{Trulens} & Relevancy & 0.368 & 0.320 & 0.289 \\
 & Faithfulness & 0.428 & 0.446 & 0.360 \\ \bottomrule
\end{tabular}}
\caption{Complete results of Trulens and ARES with \texttt{Qwen2.5-72B-Instruct} on correctness evaluation. Performance is averaged over five trials to mitigate randomness.}
\label{tab:baseline-metrics}
\end{table}

\section{Training Dynamics}
\label{sec:training_dynamics}
\begin{figure}[t]
\centering
\includegraphics[width=1\linewidth]{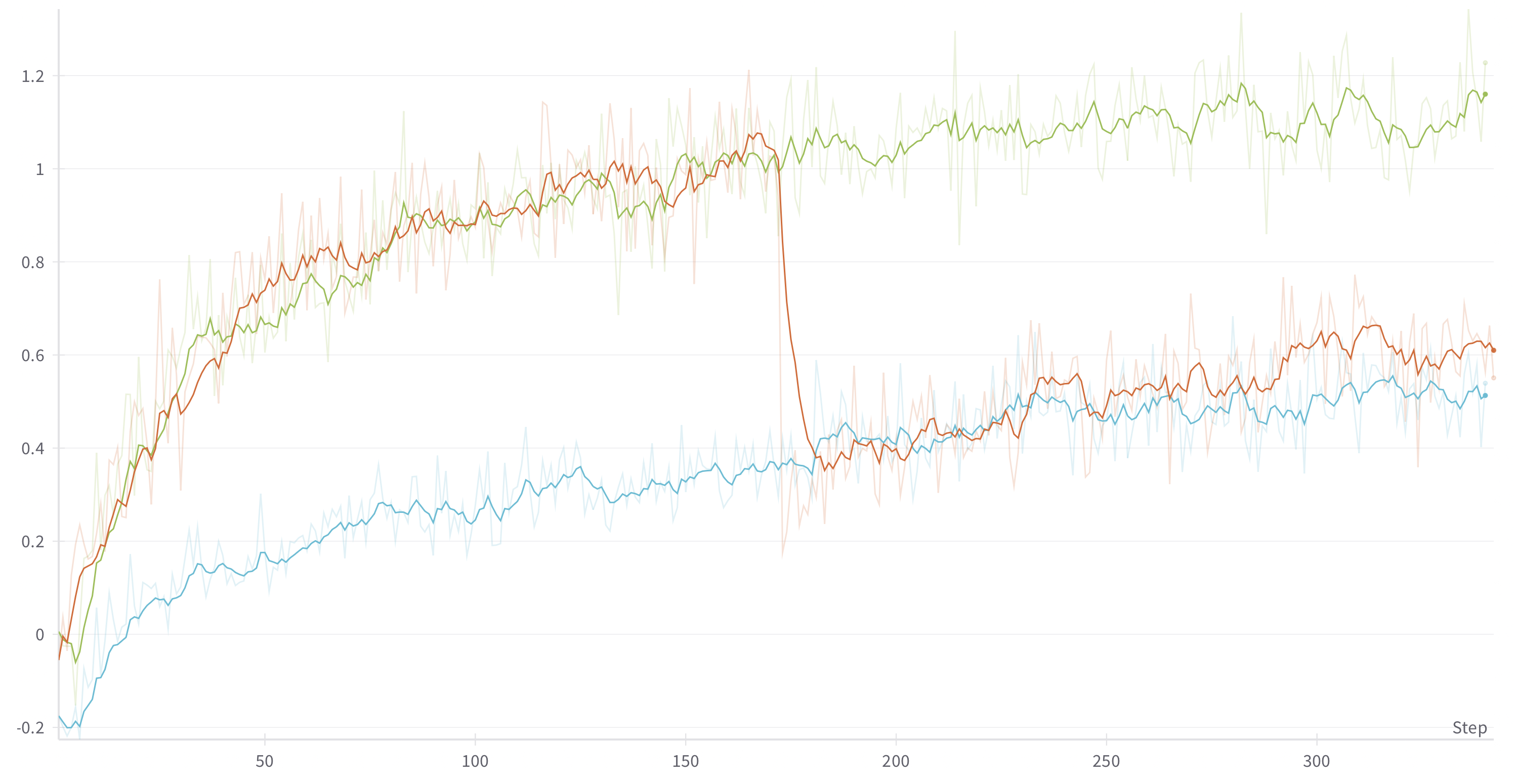}
\caption{Reward dynamics of RL training with different data configuration. The red line represents the curriculum learning settings, while the green and blue lines are for static 3 and 4 responses, respectively.  }
\label{fig:train_log}
\end{figure}

Figure~\ref{fig:train_log} presents the reward progression during RL training under different data configurations (detailed in Tab.\ref{tab:data_comparison}). The static 4-response configuration initially yields significantly lower average rewards compared to other conditions, attributable to its greater task complexity. The curriculum-based approach (red series) experiences an expected performance dip at Step 175 during the transition to 4 candidate responses, yet maintains superior average rewards over the static 4-response baseline throughout subsequent training.

\section{Prompt}
\begin{figure*}[ht]
\includegraphics[width=1\textwidth]{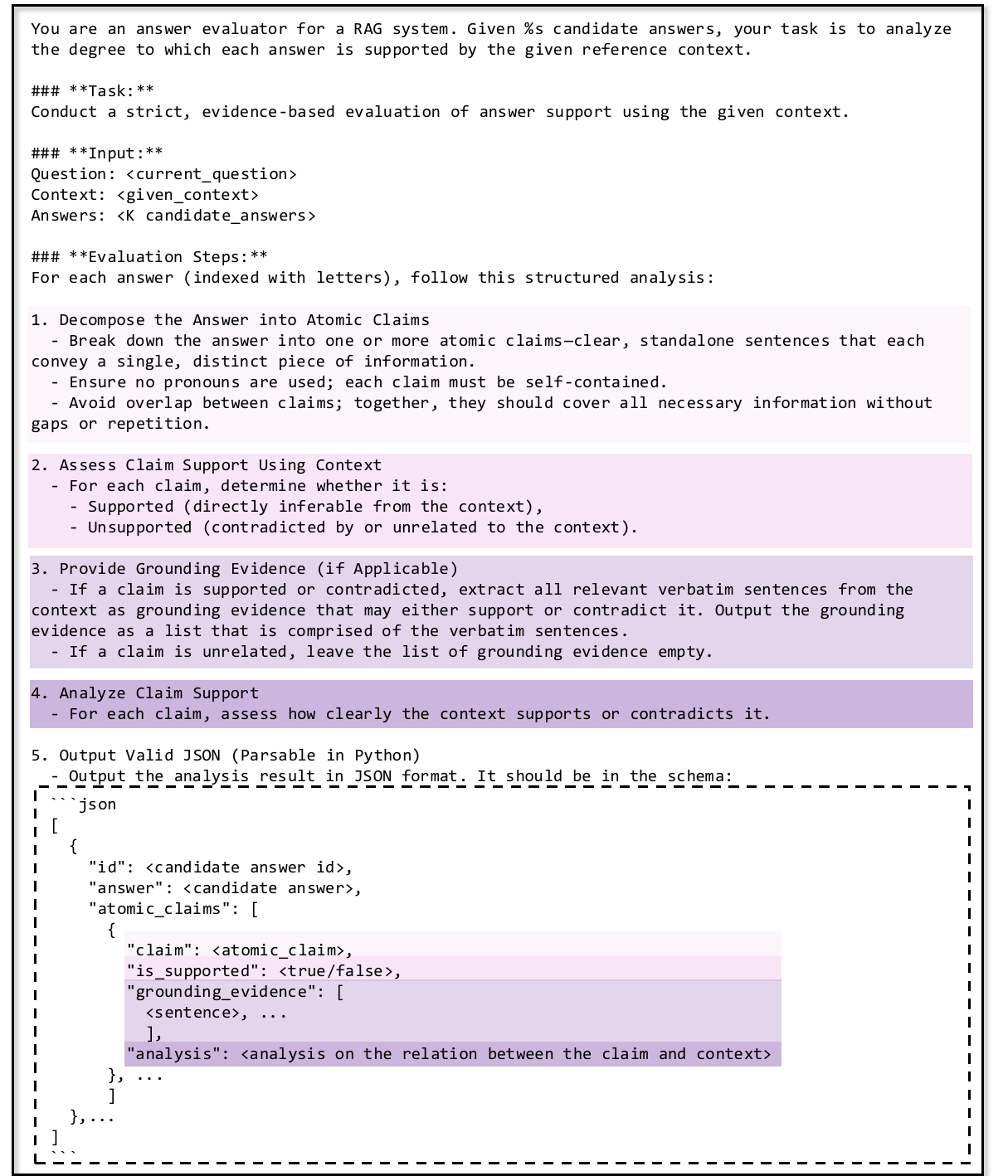}
\caption{The complete prompt used in training the evaluator. Given the current question, context, and $K$ candidate answers, the evaluator outputs a JSON-formatted string containing detailed evaluation for each candidate answer. Each evaluation follows the four key steps (highlighted in purple) to assess answer quality.}
\label{fig:prompt}
\end{figure*}

The prompt used in reinforcement learning is shown in Fig.\ref{fig:prompt}, elaborating the rules that the evaluator should conform to. Given a question, the context and $K$ candidate answers to be assessed, the model should generate a JSON-formatted output containing detailed evaluation for each candidate answer. Each answer evaluation should involve claim decomposition, claim supportive judgment, grounding evidence generation and relation analysis.

\section{Case Study}
\begin{figure*}[ht]
\includegraphics[width=1\textwidth]{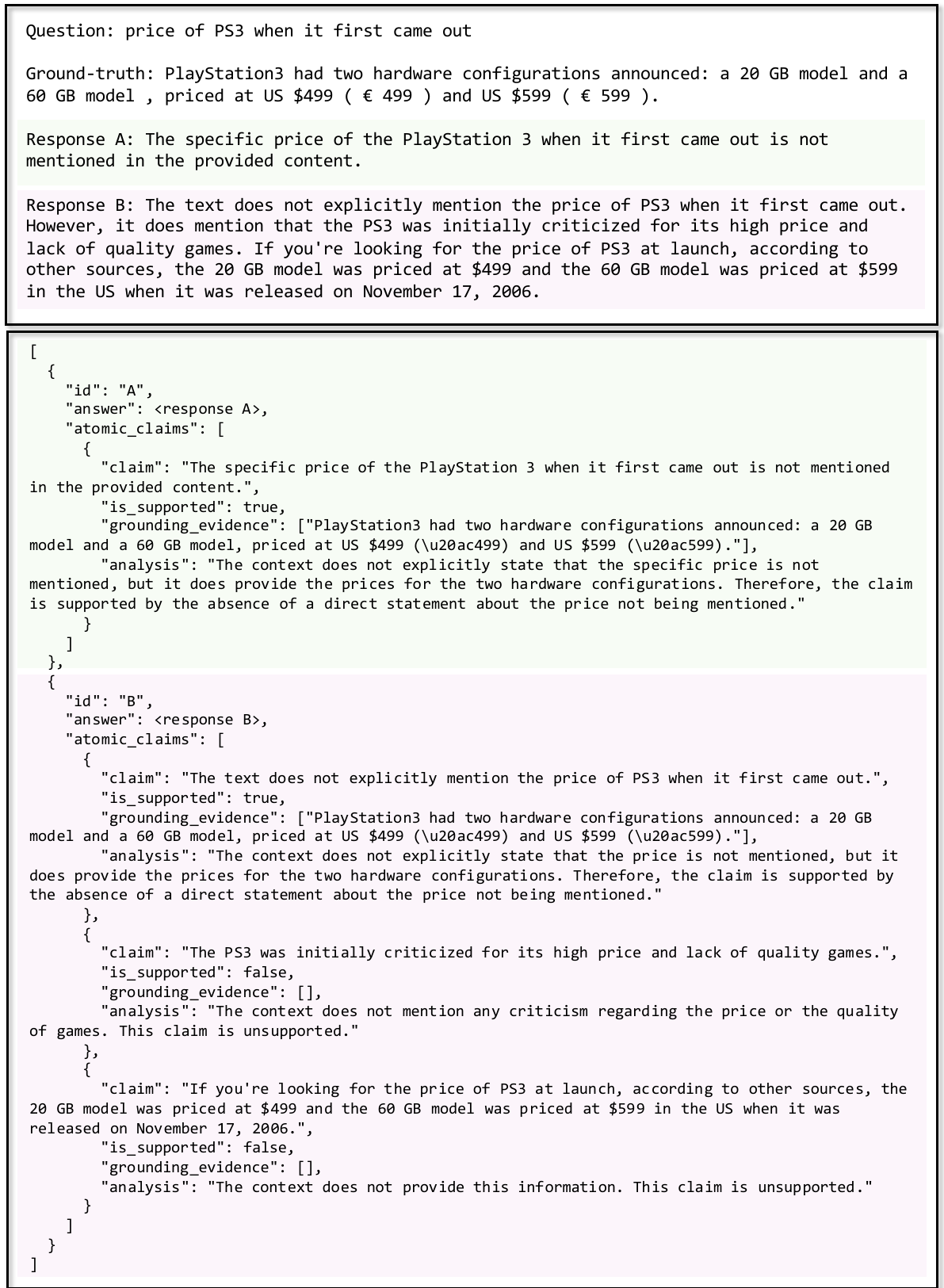}
\caption{Case study (Part-1): Human annotators judge Response B as significantly better than Response A for the given question. \textbf{RAG-ZEval w/o RL} assigns response A $1$ point, while response B $1/3$ point, misaligning with human preference. Additionally, the final claim produced by \textbf{RAG-ZEval w/o RL} is a direct copy of the original answer sentences, lacking atomic claim decomposition.}
\label{fig:case_study_norl}
\end{figure*}

\begin{figure*}[ht]
\includegraphics[width=1\textwidth]{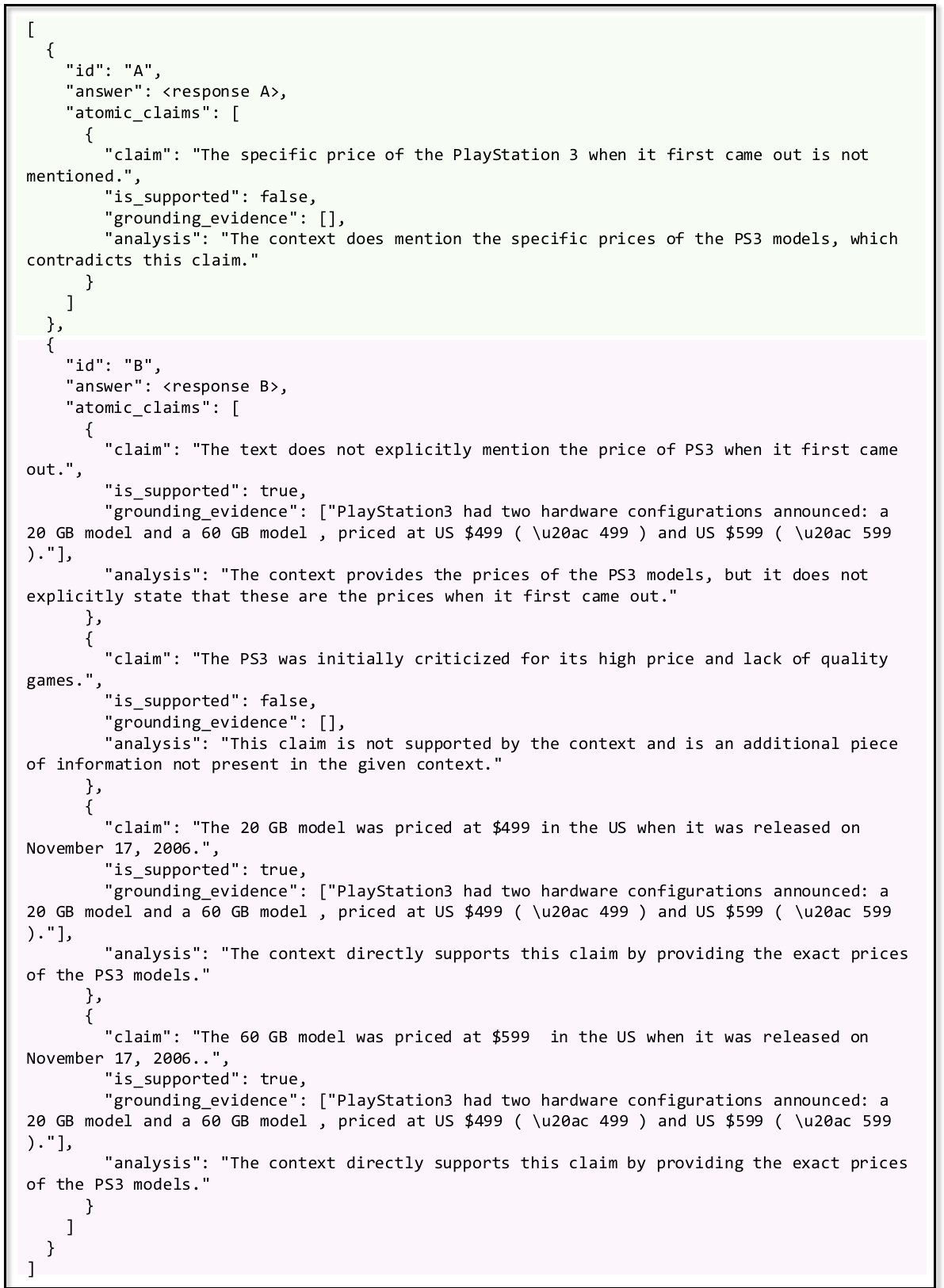}

\caption{Case study (Part-2): For the same question and responses (shown in Fig. \ref{fig:case_study_norl}), \textbf{\texttt{RAG-Zeval}} assigns response A $0$ point and response B $3/4$ point, producing a ranking consistent with human judgment.}
\label{fig:case_study_rl}
\end{figure*}

In Fig.~\ref{fig:case_study_norl} and \ref{fig:case_study_rl}, \texttt{RAG-Zeval} and its variant without RL training (\texttt{RAG-Zeval w/o RL}) generate evaluations for the same input. For this question ``\textit{price of PS3 when it first came out}'', human annotators judge response B as significantly better than response A. \texttt{RAG-Zeval w/o RL} assigns response A $1$ point, while response B $1/3$ point, resulting in an incorrect ranking $A>B$, which contradicts human preference. Additionally, its final claim is a verbatim copy of the original answer sentences, failing to perform atomic claim decomposition. In contrast, \texttt{RAG-ZEval} assigns response A $0$ point and response B $3/4$ point, producing the correct ranking that aligns with human judgment.

\centering

\begin{table*}
\small
    \centering
    \scalebox{0.85}{
    \colorbox{purple!8}{
    \begin{tabular}{@{}p{17.2cm}}
Answer the following questions based on the given passages.
\\ \\
Question: What was the purpose of designing and building the Fiat Ecobasic concept car?
\\ \\
Passage: The Fiat Ecobasic is a concept car designed by the Italian manufacturer Fiat and presented in December 1999 at the Bologna Motor Show and exhibited in March 2000 at the Geneva Motor Show. The purpose of this concept was to prove that it was possible to design and build a car capable of transporting four adults in a structure made of fully recyclable composite materials and whose production and operating costs were ultra-low.
\\ \\
Answer: The purpose of designing and building the Fiat Ecobasic concept car was to prove that it was possible to create a car that could transport four adults using fully recyclable composite materials. Additionally, the car aimed to have ultra-low production and operating costs.
\\ \\ \\
Question: When did Pope Benedict XVI become the head of the Catholic Church and sovereign of the Vatican City State, and when did he resign?
\\ \\
Passage: PPope Benedict XVI (Latin: Benedictus PP. XVI; Italian: Benedetto XVI; German: Benedikt XVI; born Joseph Aloisius Ratzinger; 16 April 1927 – 31 December 2022) was the head of the Catholic Church and sovereign of the Vatican City State from 19 April 2005 until his resignation on 28 February 2013. Benedict's election as pope occurred in the 2005 papal conclave that followed the death of Pope John Paul II. In 1981, he was appointed Prefect of the Congregation for the Doctrine of the Faith, one of the most important dicasteries of the Roman Curia. From 2002 until he was elected pope, he was also Dean of the College of Cardinals. Before becoming pope, he had been "a major figure on the Vatican stage for a quarter of a century"; he had had an influence "second to none when it came to setting church priorities and directions" as one of John Paul II's closest confidants.Benedict's writings were prolific and generally defended traditional Catholic doctrine, values, and liturgy. He was originally a liberal theologian but adopted conservative views after 1968. During his papacy, Benedict advocated a return to fundamental Christian values to counter the increased secularisation of many Western countries. He viewed relativism's denial of objective truth, and the denial of moral truths in particular, as the central problem of the 21st century. Benedict also revived several traditions, including the Tridentine Mass. He strengthened the relationship between the Catholic Church and art, promoted the use of Latin, and reintroduced traditional papal vestments, for which reason he was called "the pope of aesthetics". He was described as "the main intellectual force in the Church" since the mid-1980s.On 11 February 2013, Benedict announced his resignation, citing a "lack of strength of mind and body" due to his advanced age. His resignation was the first by a pope since Gregory XII in 1415, and the first on a pope's initiative since Celestine V in 1294. He was succeeded by Francis on 13 March 2013 and moved into the newly renovated Mater Ecclesiae Monastery in Vatican City for his retirement.
\\ \\
Answer: Pope Benedict XVI became the head of the Catholic Church and sovereign of the Vatican City State on April 19, 2005. He held this position until his resignation on February 28, 2013.
\\ \\ \\
Question: \{\texttt{question}\}
\\ \\
Passage: \{\texttt{Passage}\}
\\ \\
Answer:
\end{tabular}

}}
    \scalebox{0.85}{
    \colorbox{blue!8}{
    \begin{tabular}{@{}p{17.2cm}}
Answer the following questions.
\\ \\
Question: What was the purpose of designing and building the Fiat Ecobasic concept car?
\\ \\
Answer: The purpose of designing and building the Fiat Ecobasic concept car was to prove that it was possible to create a car that could transport four adults using fully recyclable composite materials. Additionally, the car aimed to have ultra-low production and operating costs.
\\ \\ \\
Question: When did Pope Benedict XVI become the head of the Catholic Church and sovereign of the Vatican City State, and when did he resign?
\\ \\
Answer: Pope Benedict XVI became the head of the Catholic Church and sovereign of the Vatican City State on April 19, 2005. He held this position until his resignation on February 28, 2013.
\\ \\ \\
Question: \{\texttt{question}\}
\\ \\
Answer:
\end{tabular}
}}

    \caption{Prompts used to model \textcolor{purple}{$P_{LLM} (*\mid \bm{q}, \bm{c})$} and \textcolor{blue}{$P_{LLM} (*\mid \bm{q})$},  respectively, for Context-Aware Decoding approach. The in-context examples are also sourced from Natural Question.
}
    \label{tab:appendix-prompt}
    
\end{table*}

\end{document}